\documentclass[10pt]{article} % For LaTeX2e
\usepackage[preprint]{tmlr}
\usepackage{hyperref}
\usepackage{url}

\usepackage{fontawesome}
\usepackage{graphicx, caption, subcaption, tikz, setspace, bm, mathtools, amsfonts, xfrac, etoolbox, amsthm, dutchcal, arydshln, wrapfig}
\usetikzlibrary{shapes, arrows}
\usepackage{floatrow}
\usepackage{soul}
\usepackage{subcaption}

\DeclareCaptionSubType[alph]{figure}
\usepackage[most]{tcolorbox}
\usepackage[textsize=tiny,textwidth=1.2cm, color=green!10, bordercolor=white]{todonotes}

\renewcommand{\vec}[1]{\mathrm{\mathbf{#1}}}
\renewcommand{\L}{\mathcal{L}}
\newcommand{\N}{\mathbb{N}}

\DeclarePairedDelimiter{\set}{\{}{\}}

\newtheorem{theorem}{Theorem}
\newtheorem{remark}{Remark}[theorem]
\hypersetup{colorlinks=true,
            pdfinfo={
                 Title={Assessing the Limits of In-Context Learning beyond Functions using Partially Ordered Relation},
                 CreationDate={01/10/2024--20:56:00},
                 ModDate={\the\day/\the\month/\the\year\space},
                 Creator={PdfLaTeX v3.14},
                Producer={LaTeX},
            },
            pdfkeywords={Partial Ordered Set, Sequential Learners, Large Language Models}
}
\newcommand\scalemath[2]{\scalebox{#1}{\mbox{\ensuremath{\displaystyle #2}}}}

\title{Assessing the Limits of In-Context Learning beyond Functions using Partially Ordered Relation}

\author{\name Debanjan Dutta \email debanjandutta\_r@isical.ac.in \\
      \addr Electronics and Communication Sciences Unit \\
        Indian Statistical Institute, Kolkata
      \AND
      \name Faizanuddin Ansari \email faizanuddin\_r@isical.ac.in \\
      \addr Electronics and Communication Sciences Unit \\
        Indian Statistical Institute, Kolkata
      \AND
      \name Swagatam Das \email swagatam.das@isical.ac.in\\
      \addr Electronics and Communication Sciences Unit \\
        Indian Statistical Institute, Kolkata
  }

\def\month{MM}  % Insert correct month for camera-ready version
\def\year{YYYY} % Insert correct year for camera-ready version
\def\openreview{\url{https://openreview.net/forum?id=XXXX}} % Insert correct link to OpenReview for camera-ready version

\begin{document}

\maketitle

\begin{abstract}
    Generating rational and generally accurate responses to tasks, often accompanied by example demonstrations, highlights Large Language Model's (LLM's) remarkable In-Context Learning (ICL) capabilities without requiring updates to the model's parameter space. Despite having an ongoing exploration focused on the inference from a document-level concept, its behavior in learning well-defined functions or relations in context needs a careful investigation. In this article, we present the performance of ICL on partially ordered relation by introducing the notion of inductively increasing complexity in prompts. In most cases, the saturated performance of the chosen metric indicates that while ICL offers some benefits, its effectiveness remains constrained as we increase the complexity in the prompts even in presence of sufficient demonstrative examples. The behavior is evident from our empirical findings and  has further been theoretically justified in term of its  implicit optimization process.
    The code is available \href{https://anonymous.4open.science/r/ICLonPartiallyOrderSet}{here}.\footnote{\faFileCodeO \ \  \href{https://anonymous.4open.science/r/ICLonPartiallyOrderSet}{\tiny https://anonymous.4open.science/r/ICLonPartiallyOrderSet}}  % \faGithubSquare
\end{abstract}

\section{Introduction}
% Recent works towards the advancement of \underline{L}arge \underline{L}anguage \underline{M}odels (LLMs) have paved the way for the introduction of several behavioral properties of these gigantic architectures. `\underline{I}n-\underline{c}ontext \underline{l}earning' (ICL) \cite{IclMain}, being one of such properties, has gained tremendous attention because it enables LLMs to learn tasks from the context provided by a handful of examples without parameter updates. This ability has spurred extensive research dealing with applications spanning language understanding, code generation, and mathematical reasoning \cite{Suvrit24b,Suvrit24a}.

% Partially ordered relations or posets are one of the inherent structures often found in nature. Consider a genealogical relation \emph{ancestor-of}. One provided in the Mahabharata \cite{dwaipayanmahabharata} {--} \texttt{Bashistha was the ancestor of Shaktri. Shaktri was the ancestor of Parasar, Parasar was the ancestor of Krishno-dwaipayan Bedobyas} {--} is an exampstrict linear order. \iffalse A similar genealogical chain appears in the Torah, from \texttt{Abraham} to \texttt{David}, likewise illustrating the formal properties of the \emph{ancestor-of} relation.\fi To understand the wide applicability of posets, let us consider a relation say \emph{inclusion} ($\subseteq$).  The fundamental theorem \emph{every vector space has a basis} can be proved using Zorn’s lemma \cite{zorn1935remark}. It tells a non-empty poset $P$ contains at least one maximal element provided every chain has its \emph{upper bound} in $P$. 
\underline{L}arge \underline{L}anguage \underline{M}odels (LLMs) have demonstrated remarkable capabilities in \underline{i}n-\underline{c}ontext \underline{l}earning (ICL) \citep{IclMain}, where tasks are learned from few-shot demonstrations without parameter updates. While ICL has spurred extensive research dealing with applications spanning language understanding, code generation, and mathematical reasoning \citep{Suvrit24b,Suvrit24a}, its ability to reason about relational structures—particularly partially ordered sets (posets)—remains unexplored. Posets are fundamental to mathematics and real-world hierarchies. For example, genealogical relations such as one in the Mahabharata \citet{dwaipayanmahabharata} {--} \texttt{Bashistha was the ancestor of Shaktri, Shaktri was the ancestor of Parasar, Parasar was the ancestor of Krishno-dwaipayan Bedobyas} {--} is an example of linear order. Their mathematical breadth is further illustrated by a relation say \emph{inclusion} ($\subseteq$). %The fundamental theorem \emph{every vector space has a basis} can be proved using Zorn’s lemma \cite{zorn1935remark}.  It tells a non-empty poset $P$ contains at least one maximal element provided every chain has its \emph{upper bound} in $P$. This shows posets demand reasoning about transitivity, anti-symmetry, and dependencies beyond function-like mappings.
Zorn’s lemma \citet{zorn1935remark}, one of the widely used results to prove several fundamental theorems in algebra, finds a proof involving posets. This exhibitory relevance of posets, thus demands investigation on reasoning about transitivity, anti-symmetry, and dependencies beyond function-like mappings.

Existing works on ICL have primarily explored linear functions \citep{Garg}, regular languages \citep{akyurek2024incontext}, and discrete-valued functions \citep{Satwick2024}. While these studies reveal insights into model adaptability, they overlook the challenges posed by relational reasoning, where outputs for a single image, unlike functions, may involve multiple pre-images. Along this line of study on exploring ICL's behavior on algorithms \citep{velickovic22a}, functions and even long-context NLP tasks \citep{bertsch2025context}, investigating their behavior on structured mathematical tasks remains understudied. This research-gap is critical for posets which provide a rigorous framework for evaluating reasoning in hierarchical scenarios. Yet, it is unclear whether LLMs can infer such relations through ICL, especially as task complexity scales. These challenges form the primary motivation for our research.% We aim to investigate ICL’s ability to handle relational tasks beyond one-to-one mappings by leveraging posets.

By leveraging posets, we propose a novel evaluation framework using $k$-shot $c$-complex prompts, where $k$ denotes the number of demonstrations and $c$ quantifies complexity through incremental extensions of Hasse diagrams (see Fig. \ref{fig_Hasse}). Our experiments span different open-source LLMs and models from \textsf{GPT} family, evaluating their ability to infer relations on posets: the \emph{less than} $(\mathbb{N}, <)$ and the \emph{divisibility} $(\mathbb{N}, \mid)$. The exponential growth in vertices of the Hasse diagram of poset $(2^S, \subseteq)$, for any finite set $S$ has prevented us to include them in the study. We further fine-tune models (e.g., BERT, RoBERTa) to evaluate performance when enriched with data-specific knowledge.

%%%%%%%%%%%% Old %%%%%%%%%
% \noindent
% Our key contributions are as follows: (1) We introduce a structured evaluation framework for ICL in poset relations using $k$-shot $c$-complex prompts (Section: \ref{sec:3}), (2) We conduct extensive experiments on LLMs revealing that ICL's \textbf{performance saturates} despite increasing demonstrations and complexity. While neural language models can partially encode poset relations, their reasoning is constrained by architectural limitations, particularly in capturing transitive and anti-symmetric dependencies  (Section: \ref{sec:4}).
% % revealing that ICL performance saturates despite increased demonstrations and complexity, with limitations in capturing transitivity and anti-symmetry (Section: \ref{sec:4}), 
% (3) We provide theoretical insights linking ICL’s bounded approximation capacity in its implicit optimization process, explaining the observed saturation effects (Section: \ref{sec:5}).

\noindent
Key contributions are as follows: (1) We for the first time, investigate on the \textbf{performance saturation} of ICL on posets when evaluated under the proposed framework (Section \ref{sec:3}) involving $k$-shot $c$-complex prompts. Through a series of extensive experiments (Section \ref{sec:4}), we present while LLMs can partially encode posets during ICL, their reasoning is constrained by architectural limitations, particularly in capturing transitivity and anti-symmetry. (2) We have theoretically justified this bounded approximation capacity of ICL in terms of its implicit optimization process (Section \ref{sec:5}). This becomes evident through the analysis of LLMs' output when viewed as \emph{task vectors} (Section \ref{sec:taskvectors}).

% In summary, our findings reveal that neural language models exhibit \textbf{performance saturation} despite increasing demonstrations and complexity. While they can partially encode poset relations, their reasoning is constrained by architectural limitations, particularly in capturing transitive and anti-symmetric dependencies.

% \subsection{Revisiting the ICL Landscape}
% To contextualize our findings within the broader ICL literature, we have discussed two key related areas of study. i) The empirical works addressing discrete function along with long-context NLP tasks, and ii) The relationship between the learning paradigm of ICL and gradient-based optimization. Detailed discussion is provided in the Appendix \ref{} due to space limitations.

\section{Revisiting the ICL Landscape}
\label{sec:2}
% Understanding ICL in LLMs has been extensively studied, with research focusing on theoretical foundations and the challenges of extending ICL to relational structures. 
Recent studies have established a strong theoretical basis for ICL by analyzing its connection to gradient-based optimization. Building on \citet{Suvrit24b}'s work on linear Transformers performing in-context gradient descent, \citet{Suvrit24a} demonstrated similar behavior in non-linear Transformers under specific activation conditions.
% \citet{Suvrit24b} demonstrated that linear Transformers can perform gradient descent in-context, effectively learning linear functions through parameterized attention mechanisms. Extending this work, \citet{Suvrit24a} showed that non-linear Transformers exhibit similar behavior under specific conditions where activation functions align with the data distribution, thereby generalizing ICL to more complex function classes. 
Furthermore, \citet{dai-etal-2023-gpt} empirically validated the duality between Transformer attention and gradient descent, showing in-context updates mimic implicit fine-tuning, particularly when leveraging momentum-based attention.

% \citet{Satwick2024} examined frozen-weight GPT-2 models on discrete-valued functions, such as the \emph{nearest neighbor} task, and found non-trivial performance, suggesting that LLMs encode structural priors useful for learning certain function classes. Other works, such as \citet{akyurek2024incontext}, focused on regular language classes, highlighting cases where Transformer-based models struggle to generalize. Despite these advancements, a critical limitation remains: most prior studies assume function-based mappings, leaving relational reasoning underexplored. 
In the context of analyzing ICL on functions from discrete space: \citet{Satwick2024} found frozen-weight GPT-2 models showed non-trivial performance on discrete functions like nearest neighbor. \citet{akyurek2024incontext} highlighted in-context (regular) language learning excels via implicit implementation of $n$-gram statistics. \citet{bertsch2025context} noted diminishing gains in long-context ICL performance in  traditional NLP tasks, suggesting fundamental limitations in contextual inference. Moreover, \citet{akyurek2023what} and \citet{guo2024how} demonstrated that ICL effectiveness is highly dependent on the function properties, prompting the question of how LLMs perform when dealing with relations and partial functions rather than well-structured functions. By analyzing poset, we extend ICL research into relational reasoning, filling up a crucial gap in understanding how language models infer dependencies beyond standard function learning. 

\section{Background and Setup for ICL}
\label{sec:3}
\subsection{Partially Ordered Set \& Hasse Diagram}
    \textbf{Poset}: 
    On a set $S$, a relation $\preceq$ is said to produce a partial-ordered set or poset $(S, \preceq)$, if it follows:
    \begin{description}\vspace{-5pt}
        \itemsep-2pt
        \item[ ]\emph{Reflexivity} For every element $s \in S$, $s \preceq s$,
        \item[ ]\emph{Anti-symmetry} For any two elements $s_1, s_2 \in S$, if $s_1 \preceq s_2$ and $s_2 \preceq s_1$, $s_1 = s_2$ and
        \item[ ]\emph{Transitivity} For three elements $s_1, s_2$ and $s_3 \in S$ if $s_1 \preceq s_2$ and $s_2 \preceq s_3$, $s_1 \preceq s_3$.\vspace{-5pt}
    \end{description}
    
    A poset, in absence of reflexivity property, denoted by $(S, \prec)$ is called linear order if any two distinct elements $i, j\in S$ are comparable i.e. either $i\prec j \text{ or } j \prec i$. 
    % A linear order is further categorized as dense if for any two $i, k \in S \text{ and } i \prec k$, there exists a $j \in S$ such that $i \prec j \prec k$  where $|S| \ge 2$. 
    Our work comprises study on two posets {--} the \emph{less than} relation on the set of natural numbers $(\N, <)$, representative of any linear order, and the \emph{divides} relation on the set of natural numbers $(\N, \mid)$, representing any other arbitrary poset.

    \emph{Observation} \hypertarget{obs}{\emph{1}}. The linear-order on numerical domain such as $(\N, <)$ can be viewed as length-dependent lexicography order. Suppose $\Sigma_\N = \set{1, 2, \ldots, 9} \cup \set{0}$ denotes the alphabet of $\N$. Then a natural word $x=a_1a_2\ldots a_n$ of length $n$ belongs to $\Sigma_\N^n$ if and only if $a_i\in\Sigma_\N$. Suppose the lexicography order on $\Sigma_\N$ is i) the linear order $(0,1,2,\dots, 9)$ and ii) $x$ precedes $x^\prime$ for any natural word $x \in \Sigma_\N^i \text{ and } x^\prime \in \Sigma_\N^{i+k}$ where $k > 0$. Then length-dependent lexicography order on $\Sigma_\N$ is the linear order $(\N, <)$.
    
    \begin{wrapfigure}{r}{0.35\textwidth} %only to get both on one page
    % \vspace*{-1em}
      \begin{subfigure}{0.45\columnwidth} \centering
      \begin{tikzpicture}[scale=.45, > = stealth]
            \node (four) at (0,4.8) {$4$};
            \node (three) at (0,3.2) {$3$};
            \node (two) at (0,1.6) {$2$};
            \node (one) at (0,0) {$1$};
            \path[->] (one) edge (two);
            \path[->] (two) edge (three);
            \path[->] (three) edge (four);
      \end{tikzpicture}
      \caption{Hasse representing $(\set{1,\dots,4}, <)$ or the linear order $(1,2,3,4)$.}
      \end{subfigure}\hspace*{\fill}
      \begin{subfigure}{0.5\columnwidth}\centering
          \begin{tikzpicture}[scale=.45, > = stealth]
              \node (six) at (0,2) {$6$};
              \node (five) at (2,0) {$5$};
              \node (four) at (-2,2) {$4$};
              \node (three) at (0,0) {$3$};
              \node (two) at (-2,0) {$2$};
              \node (one) at (0,-2) {$1$};
              \path[->] (one) edge (two);
              \path[->] (two) edge (four);
              \path[->] (one) edge (three);
              \path[->] (one) edge (five);
              \path[->] (two) edge (six);
              \path[->] (three) edge (six);
              \path (one) edge [loop below] (one);
              \path (two) edge [loop left] (two);
              \path (three) edge [loop right] (three);
              \path (four) edge [loop above] (four);
              \path (five) edge [loop right] (five);
              \path (six) edge [loop above] (six);
          \end{tikzpicture}
          \caption{Hasse representing $(\set{1,\dots,6}, \mid)$.}
      \end{subfigure}
      \caption{Examples of Hasse diagrams.\vspace*{-1em}}
      \label{fig_Hasse}
    \end{wrapfigure}
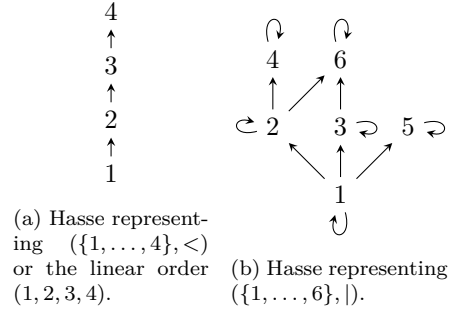
    \textbf{Hasse Diagram}:
    Given a poset $(S, \preceq)$, a Hasse diagram (e.g. Fig: \ref{fig_Hasse}) is a directed acyclic graph (DAG) $H = (S, E)$ where $E \subset S \times S$ represents the directed edge $(a, b)$ such that $a \preceq b$ and there exists no $c$ such that $a \preceq c \preceq b$ when $a \ne b$ and self-loop in case of $a=b$ depending on reflexivity (or, irreflexivity). The Hasse diagram of the relation \emph{less than} is a directed path. Suppose not, then there exist at least two paths say from $a$ to $b_1$ and $a$ to $b_2$ such that $b_1$ and $b_2$ are not connected either in direction $b_1$ to $b_2$ or vice-versa. Then, by definition of Hasse $b_1$ and $b_2$ are incomparable contradicting the basic rule that with any two different integers, one is smaller than the other. The Hasse diagram of the relation \emph{divides}, on the other hand, can contain such $b_1$ and $b_2$, representing any arbitrary DAG.
    
    \subsection{Formalizing $k$-Shot $c$-Complex Prompt}
    \label{subsec:ICL}
    % A simple yet effective mathematical framework proposed in \citet{Garg} is well-equipped to define a $k$-shot prompt $P_k$ and the ingredients to define ICL. A neural language model $\L$ can generate a response $\L(P_k)$ to a $k$-shot prompt $P_k = \sequence{\vec{x}_1, \vec{y}_1, \dots, \vec{x}_k, \vec{y}_k, \vec{x}_{k+1}}$, where $\vec{y}_i \in f(\vec{x}_i)$ and $f : D \times C$ is any relation for some arbitrary domain $D$ and co-domain $C$. We say, a $k$-shot prompt $P_k$ has successfully provided a context to $\L$ or $\L$ is learning in-context if $\L(P_k) \in f(\vec{x}_{k+1})$ or loosely, $\operatorname{dist}(\L(P_k), \vec{y}_{k+1}) < \epsilon$ for a distance metric $\operatorname{dist}$. % : \mathcal{A}\times\mathcal{A} \to \mathbb{R}$ where $\mathcal{A}$ is any arbitrary domain and $\vec{y}_{k+1}\in f(\vec{x}_{k+1})$. 

    \noindent
    % \textbf{Prompts to explore ICL}: 
    Let's revisit the existing definition of ICL \citep{Garg} in relevance to our experimental context. Assuming $\Sigma$ represents the alphabets in the poset, the grammar producing a prompt $P$ has been provided as Backus–Naur form in Figure \ref{fig:gram1}.\ %a prompt $P$ can be generated as described in Figure \ref{fig:gram1}.
    A prompt is a concatenation of instruction ($I$), examples ($E$), and task ($T$). The instruction provides information about the alphabets in the poset and definition of anti-symmetry, transitivity, and/or reflexivity depending on the nature of the relation. The task describes the output format and test point(s). %includes a description of the output format and a test point.
    
    \begin{wrapfigure}[13]{r}{0.5\textwidth}
    \small \vspace*{-2em}
        \begin{align*}
            & S \to I E T\\
            & I \to \Sigma R\\
            & R \to Reflexivity \ R^\prime \mid Irreflexivity \ R^\prime\\
            & R^\prime \to Anti\text{-}Symmetry \ Transitivity\\
            & E \to a \preceq b, \ E \mid a \preceq b \text{ \hspace{7mm} for } a, b \in \Sigma^+ \text{ and } a \preceq b\\
            & T \to D T^\prime\\
            & D \to \text{Output True or False to the following case(s) }\\
            & T^\prime \to a \preceq b, T^\prime \mid a \preceq b \text{ \hspace{1cm} for } a, b \in \Sigma^+. % a \preceq b, \ T^\prime \mid
        \end{align*}\vspace*{-2em}
        \caption{Grammar for production of a prompt. Illustrative prompts have been provided in Appendix \ref{ssec:A1}.\vspace*{-1em}}
        \label{fig:gram1}
    \end{wrapfigure}
    A prompt is generally termed as $k$-shot prompt $P_k$ if the non-terminal $E$ has expanded $k$ times. Further, we call $P_k$ is \emph{minimal} if the examples in $P_k$ are the pairs $a \preceq b$ such that $(a,b)$ are edges in the respective Hasse diagram. This can be considered being aligned with the concept of a \emph{teaching sequence} \citet{goldman1995complexity}. The approach constructs a sequence $\set{x, f(x)}$ (here, a relation) capable of uniquely identifying a function $f$ within a given function class. Minimal prompts were carefully designed to ensure interpretability and compactness while retaining task specificity. Thus, the examples combined with the instruction segment of a prompt $P_k$ can provide information about the entire poset $(\set{1,2,\dots,k}, \preceq)$.

    % \textbf{Inductive Complexity in Prompts}:  Motivated by one of the elegant yet simple researches~\cite{Xu2024HallucinationII}, showing there exists a language that cannot be produced by any iterative learning algorithms such as the neural language models and thereby making false response or \emph{hallucination} inevitable, we extend the experiments to understand at what extent ICL can remedy this problem. Because a sufficiently trained neural language model cannot generate at least one language, it may sound intuitive to ask {--} g
    \citet{Xu2024HallucinationII} proved existence of a (formal) language that cannot be produced by any iterative learning algorithms such as neural language models. We extend the experiments to understand at what extent ICL can remedy this problem. Given any existing language model $\L$, a problem class $Q$ and a problem $q \in Q$, \emph{bad} for $\L$; we ask at what extent the actual answer of $q$ deviates from the answer generated by $\L$. % However, through carefully designed experiments, we demonstrate while ICL offers improvements, it remains constrained by certain limitations.
    
    Given some knowledge about a problem class $Q$, suppose $Q$ is enumerable in terms of its increasing \emph{complexity} with respect to a problem $q^\prime \in Q$ and say $q \in Q$ is the $i$\textsuperscript{th} instance for which a neural language model $\L$ fails to generate correct answer. The term complexity here denotes the difficulty in accurately representing $q$, given the existing knowledge about class $Q$, the known problem $q^\prime$, and the inherent knowledge of $\L$. Then intuitively, the $i+1, \ldots $ problems would not only be mistaken by $\L$ but the mistake would get \emph{severe} as we go right. Despite being restricted to poset, with a suitable definition of complexity, the experiment can be reproduced analogously for suitable $Q$. We discovered this intuition does not hold true as ICL induces different outcomes.

    % For defining problems of comparable complexity, we further modify the production of $T^\prime$ in Figure~\ref{fig:gram1} as $T^\prime \to a \preceq b, \ T^\prime \mid a \preceq b \text{ for } a, b \in \Sigma^+$ and define a $k$-shot $c$-complex prompt $P_{k,c}$ as follows $(k, c \ge 1)$.
    Now we define a $k$-shot $c$-complex prompt $P_{k,c}$ is a minimal $k$-shot prompt $P_k$ on a poset $(\set{1,2, \ldots, k}, \preceq)$ which provides task $T^\prime$ that is $c$-complex, meaning $T^\prime$ provides evaluation points $a \preceq b$ such that the edge $(a, b)$ or $(b,a)$ is not present in the Hasse diagram of the poset and at least one of $a, b \in \set{k+1, k+2, \ldots, k+c}$.% See Appendix \ref{example1} for illustrative examples.\footnote{Prompts can be reproduced using the code referred earlier.}
    The rationale behind modifying the test point production rule $T^\prime$ is {--} had $T^\prime$ been defined as $a \preceq b$, the evaluation of $\mathcal{L}(P_{k,c})$ would lack robustness.  Therefore, a sequence of $c$-complex prompts, $P_{\to, c}$\footnote{The notation denotes the sequence $\set{P_{k,c}}_{k =\set{1, 2, \ldots}}$}, enhances the understanding of the provided data and is expected to yield improved performance. On the other hand, a sequence of $k$-shot prompts $P_{k, \to}$ progressively increases the task's difficulty.

    \subsection{Experimental Setup \& Evaluation Metric}
    During this experiment, we chose to keep the cardinality of evaluation points, $|P_{k,c}[T^\prime]|$, as $50$ and $30$ for the tasks \emph{less than} and \emph{divides} respectively. For example, consider the prompt $P_{20, 10}$. There are exactly $19$ demonstrations given by $1<2, 2<3, \dots, 19<20$ and $50$ random evaluation points $(a,b)$ from $\set{1\ldots,30}\times\set{21,\ldots, 30}$ or the reversed order such that the pairs are not present in demonstration. If $a\preceq b$ is the $i$\textsuperscript{th} evaluation point, $P_{k,c}[T^\prime][i]$, we say its target value $V(P_{k,c}[T^\prime][i])$ is true if there is a path from $a$ to $b$ in the Hasse diagram or false, otherwise. Then the accuracy of a generated response corresponding to a prompt $P_{k,c}$ is measured by 
\[
\scalemath{0.8}{
    \frac{1}{|P_{k,c}[T^\prime]|}\sum_i \mathbb{I}\left(\L(P_{k,c})[i], V(P_{k,c}[T^\prime][i])\right),} \vspace{-5pt}
\] 
    where, $\mathbb{I}(a, b)$ yields $1$ when $a = b$ and $0$ otherwise. Similarly, the mean cumulative accuracy, say, for a sequence of $c$-complex prompts, is given by
\[ \scalemath{0.8}{
    \frac{1}{\sum\limits_{\substack{\mathcal{c} \le c\\ \text{ for all } k}}|P_{k,\mathcal{c}}[T^\prime]|}\sum_{\substack{\mathcal{c} \le c\\ \text{ for all } k}}\sum_i \mathbb{I}\left(\L(P_{k, \mathcal{c}})[i], V(P_{k, \mathcal{c}}[T^\prime][i])\right).} \vspace{-5pt}
\] 
    %For example, suppose there is a sequence of $3$-complex prompts $P_{1,3}, \ldots, P_{10, 3}$. Then, the mean cumulative accuracy calculated on the evaluation points of this sequence is the ratio between the sum of numbers of the correct answers produced by the language models corresponding to the prompts $P_{1,1}, \ldots P_{10, 1}, P_{1,2}, \ldots P_{10, 2}, P_{1,3}, \ldots P_{10, 3}$ to the total number of evaluation points present in all these prompts.
    % Similar evaluation metric can also be incorporated for varying demonstrative examples present in a sequence of prompts keeping the complexity parameter same on average. \\
     Similar evaluation metric can also be incorporated for varying demonstrative examples keeping the complexity parameter same on average. \\
    % \todo{does this make sense?}
    \noindent
    \textbf{Models and Plots}: We have employed five decoder based models, viz \textsf{Gemma2}, \textsf{Llama3}, \textsf{Mathstral}, \textsf{Qwen2-math}, \textsf{Phi3} and two prevalent models from \textsf{GPT} family viz \textsf{GPT-3.5-Turbo} and \textsf{GPT-4.o-mini} for ICL based experiments while for fine-tuning we have employed two encoder-only transformer architectures, BERT and RoBERTa and an auto-regressive architecture XLNet, an improvement over BERT (for model details see Appendix Table \ref{tab:Models}). Despite belonging to decoder-based category, the well-known architectures exhibit varying performance on same dataset, hence selecting them to assess their efficiency. We have plotted mean cumulative accuracy for the sequence $\set{\L(P_{k,c})}_{k =\set{1, 2, \ldots, K}}$ against varying complexities, for some $K$. It denotes that we are plotting mean cumulative accuracy for $\set{\L(P_{k,1})}_{k =\set{1, \ldots, K}}, \set{\L(P_{k,2})}_{k =\set{1, \ldots, K}}, \ldots$. This implies maintaining similar number of demonstrations on average, we are increasing complexity as we go right.    
\section{In-Context Learning on Poset}
We present the first analysis of ICL for poset reasoning through systematic evaluation across: (1) linear orders $(\mathbb{N}, <)$ , (2) its binary variant  $(\{0,1\}^*, <)$ (testing length-aware lexicographic reasoning), and (3) divisibility posets $(\mathbb{N}, |)$ (challenging DAG-based transitive inference). 
% While exponentially growing posets (e.g., $(2^S, \subseteq)$ were excluded due to computational intractability, our framework establish core benchmarks for hierarchical reasoning in LLMs. 
Performance of open-source language models on the \emph{linear order} and \emph{divisibility} posets in subsections \ref{sec:lo} and \ref{sec:div} respectively, keeping analysis on the \textsf{GPT} models separate in subsection \ref{sec:gpt}.
\label{sec:4}

        \subsection{Linear Order}
        \label{sec:lo}
\begin{figure}
            %\vspace*{-3mm}
          \ffigbox{%
            \begin{subfloatrow}
              \fcapside[\FBwidth]
                {\caption{}\label{fig:1a}}%
                {\includegraphics[trim={4.5mm 0 0 0mm}, clip, width=0.3\textwidth]{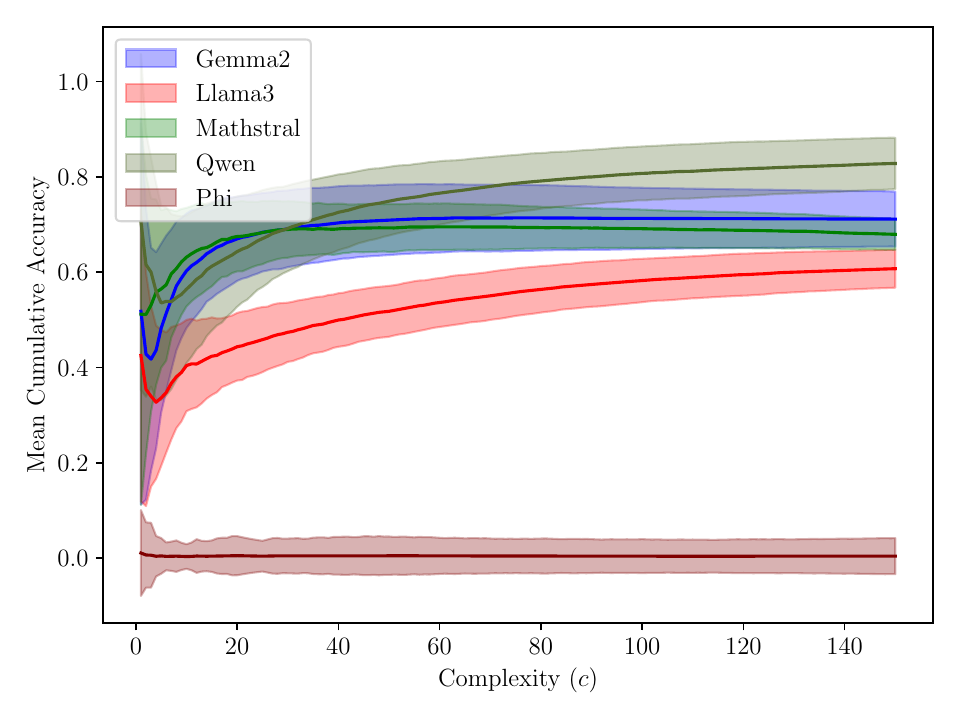}}%
           \end{subfloatrow}\hspace{\fill}
           \begin{subfloatrow}
              \fcapside[\FBwidth]
                {\caption{}\label{fig:1b}}%
                {\includegraphics[trim={4.5mm 0 0 0mm}, clip, width=0.3\textwidth]{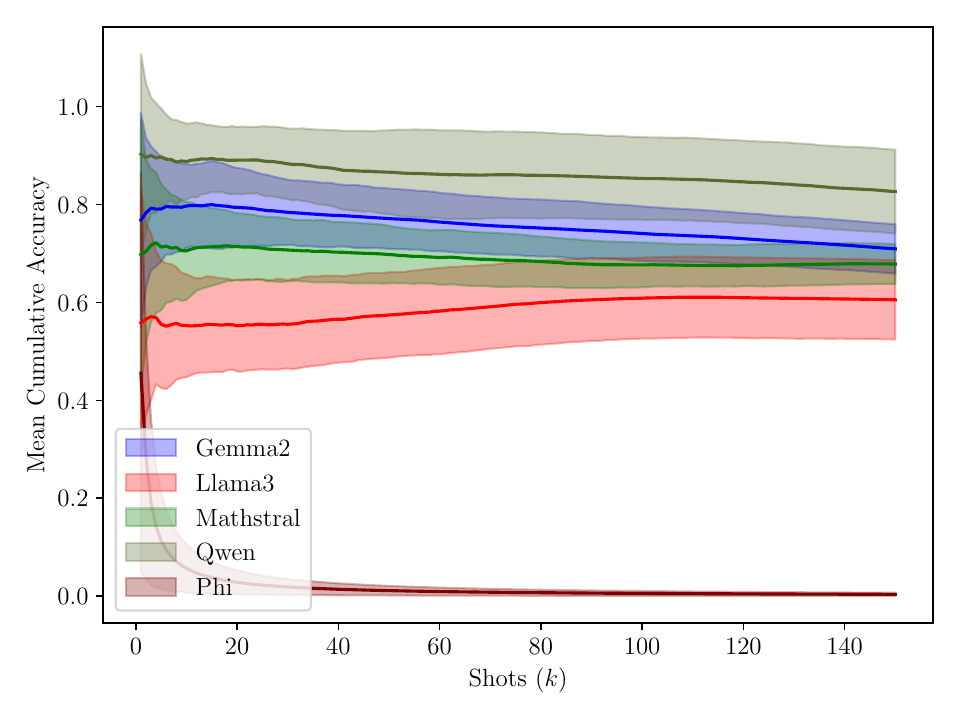}}%
            \end{subfloatrow}\hspace{\fill}
            \begin{subfloatrow}
              \fcapside[\FBwidth]
                {\caption{}\label{fig:1c}}%
                {\includegraphics[trim={4.5mm 0 0 0mm}, clip, width=0.3\textwidth]{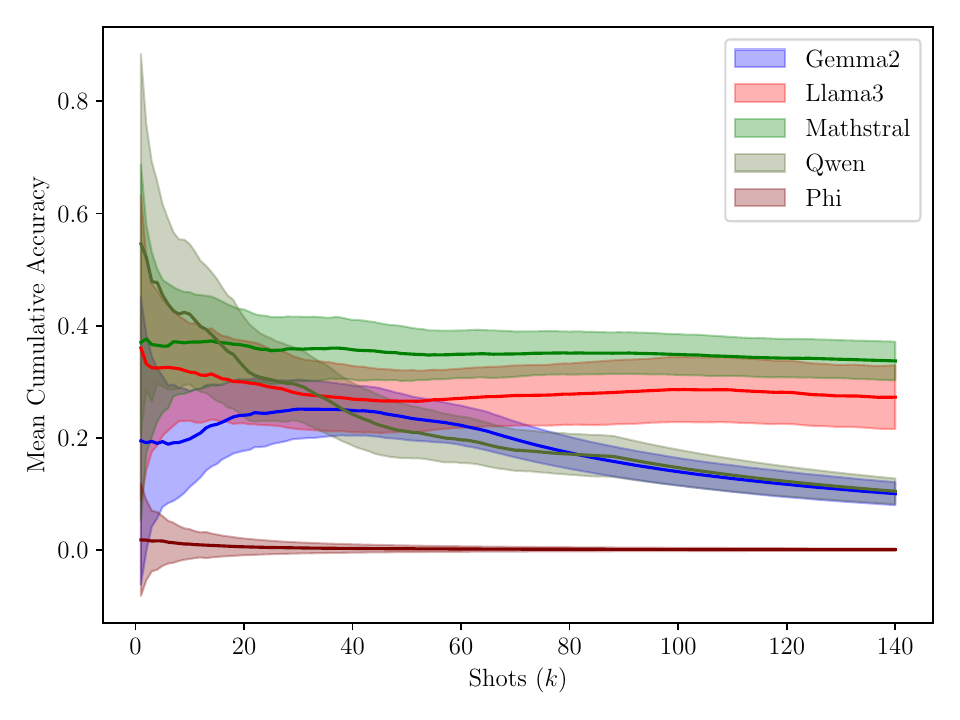}}%
            \end{subfloatrow}
            }
          {\caption{Subfigure \ref{fig:1a} describes the mean cumulative accuracy plot for the sequence $\set{\L(P_{k,c})}_{k =\set{1, 2, \ldots, 150}}$ against varying complexities. Subfigure \ref{fig:1b} describes the same metric for $\set{\L(P_{k,c})}_{c =\set{1, 2, \ldots, 150}}$ against varying shots. Subfigure \ref{fig:1c} represents the mean cumulative accuracy curve for linear order $(\set{0,1}^*, <)$.\vspace*{-1em}}
        \label{fig:1}}
        \end{figure}

         We have experimented with ICL on linear order using posets $(\N, <)$ and $(\set{0,1}^*, <)$, on a collection of $k$-shot, $c$-complex prompts, $k,c \le 150$. Figure \ref{fig:1a}  and \ref{fig:1b} depict the performance of the language models on poset $(\N, <)$. Irrespective of the size of language models, majority of them (except Qwen) failed to learn the pattern successfully. 

         Figure \ref{fig:1a} indicates that maintaining a fixed knowledge representation, on average, does not impact ICL capabilities as complexity increases, and it remains unchanged over sufficiently complex regions for models {--} \textsf{Gemma2, Llama3} and \textsf{Mathstral}, though \textsf{Qwen2-math} has shown non-trivial performance increase. Similarly, Figure \ref{fig:1b} demonstrates that increasing knowledge has no observable effect on learning performance. In fact, we see either stagnant behavior (after $115$ in case of \textsf{Llama3} and \textsf{Mathstral}) or a mildly decaying accuracy (in case of \textsf{Gemma2} and \textsf{Qwen2-math}).

         We performed similar experiments on its binary variant. The motivation behind reducing the alphabet from $\set{0,1,\ldots,9}$ to $\set{0, 1}$ is to validate observation \hyperlink{obs}{1}. By using length-dependent lexicography, achieved by converting integers to their binary equivalents, we observed certain key characteristic while analyzing the sequence $\set{\L(P_{k,c})}_{c=\set{1,2,\ldots,150}}$ against varying shots, as shown in Figure \ref{fig:1c}. Although both \textsf{Llama3} and \textsf{Mathstral} reach saturation despite having a poor performance compared to $(\N, <)$ , both \textsf{Gemma2} and \textsf{Qwen2-math} demonstrate a prominent decline in mean cumulative accuracy, which appeared milder in the case of $(\N, <)$. However, a sharp performance gap across models for tasks $(\N, <)$ and $(\set{0,1}^*, <)$ (Fig. \ref{fig:1b}, \ref{fig:1c}) underscores their failure to generalize lexicography.
         The trivial performance of \textsf{Phi3} confirms its incapability of processing minimal prompts for linear orders when we maintain identical the experimental setup for each model.
         
    \subsection{Division}
    \label{sec:div}
    In contrast to linear order, the poset $(\N, \mid)$ represents a more general class of partial order in the sense that the Hasse diagram of $(\N, \mid)$ is a general directed acyclic graph having a source vertex $1$ and an integer $n = \prod p_i^{q_i}$ for primes $p_i$ such that $q_i > 0$ has edges from $\sfrac{n}{p_i}$. In our experiment, as we progressed to 150-shot 150-complex prompts, the ICL is expected to construct incrementally a Hasse diagram of height at most $1 + \lfloor \log_{2}300 \rfloor$.
\begin{figure}
    % \vspace*{-3mm}
  \ffigbox{%
    \begin{subfloatrow}
      \fcapside[\FBwidth]
        {\caption{}\label{fig:2a}}%
        {\includegraphics[trim={4.5mm 0 0 0mm}, clip, width=0.45\textwidth]{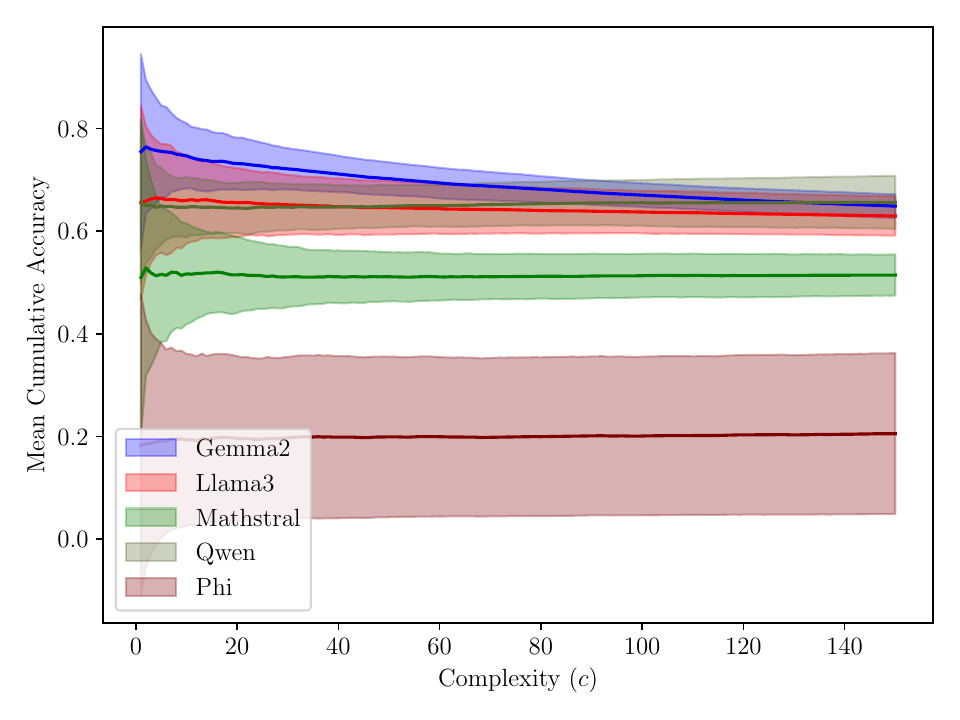}}%
   \end{subfloatrow}\hspace*{\fill}
   \begin{subfloatrow}
      \fcapside[\FBwidth]
        {\caption{}\label{fig:2b}}%
        {\includegraphics[trim={4.5mm 0 0 0mm}, clip, width=0.45\textwidth]{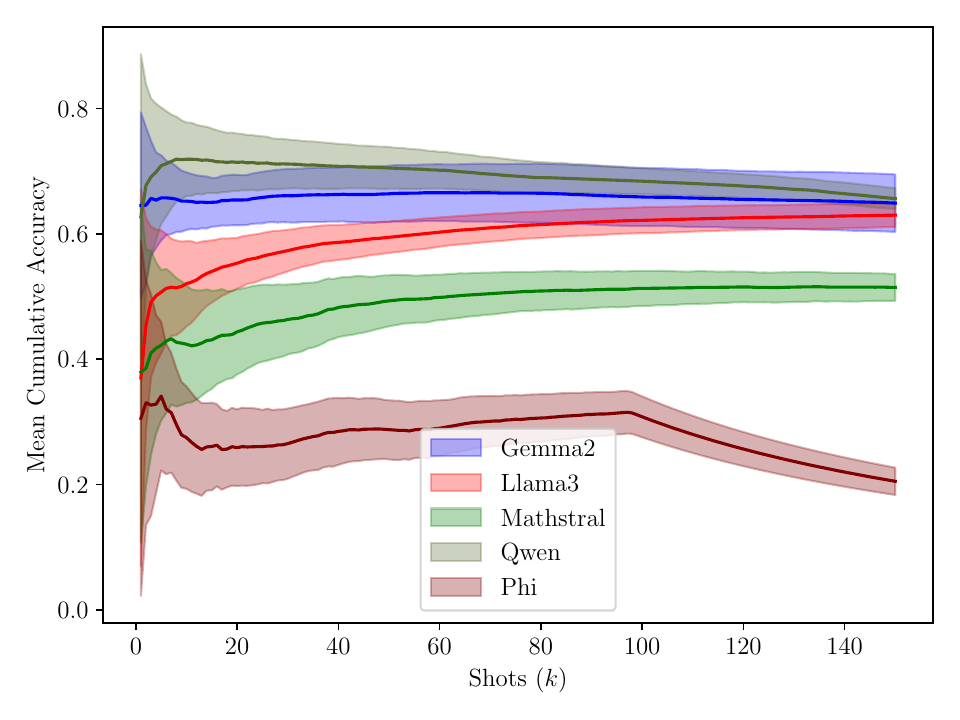}}%
    \end{subfloatrow}\\[-2mm]
    }
  {\caption{Subfigure \ref{fig:2a} depicts the mean cumulative accuracy plot for the sequence $\set{\L(P_{k,c})}_{k =\set{1, 2, \ldots, 150}}$ against varying complexities. Subfigure \ref{fig:2b} similarly describes the same metric for $\set{\L(P_{k,c})}_{c =\set{1, 2, \ldots, 150}}$ against varying shots.\vspace*{-1em}}
        \label{fig:2}}
\end{figure}
    
    Considering $|P_{k,c}[T^\prime]| = 30$, except some pathological cases, Figure \ref{fig:2a} and \ref{fig:2b} analogously describe the mean cumulative accuracy plot for the sequence $\set{\L(P_{k,c})}_{k =\set{1, 2, \ldots, 150}}$ against varying complexities and for the sequence $\set{\L(P_{k,c})}_{c =\set{1, 2, \ldots, 150}}$ against varying shots, respectively. The former describes when the knowledge remains constant on average, increasing complexity has no observable effect on ICL for \textsf{Llama3}, \textsf{Mathstral}, \textsf{Qwen2-math} and \textsf{Phi3}. However, \textsf{Gemma2} demonstrates a declining learning trend under the same conditions. Interestingly, as knowledge accumulation increases, as illustrated in Figure \ref{fig:2b}, ICL yields a notable performance enhancement in the \textsf{Llama3}, \textsf{Mathstral} and \textsf{Phi3} models, though only within a limited region. The saturation in learning can be observed in both \textsf{Mathstral} and \textsf{Gemma2}. The abrupt descent in performance for \textsf{Phi3} is due to its inability to encode context with a window of size $4096$, which we keep identical for evaluation across all the open-source models.\\
    \textbf{On fine-tuning.} For both the posets, performance on fine-tuning have been presented in Figure \ref{fig:finetune} (in Appendix \ref{ssec:A1}). We found while all three models eventually attains saturation, \iffalse under the same evaluation metric,\fi RoBERTa adapts more efficently than BERT and XL-Net.
\begin{figure}[!ht]
    % \vspace*{-3mm}
  \ffigbox{%
    \begin{subfloatrow}
      \fcapside[\FBwidth]
        {\caption{}\label{fig:3a}}%
        {\includegraphics[trim={4.5mm 0 0 0mm}, clip, width=0.45\textwidth]{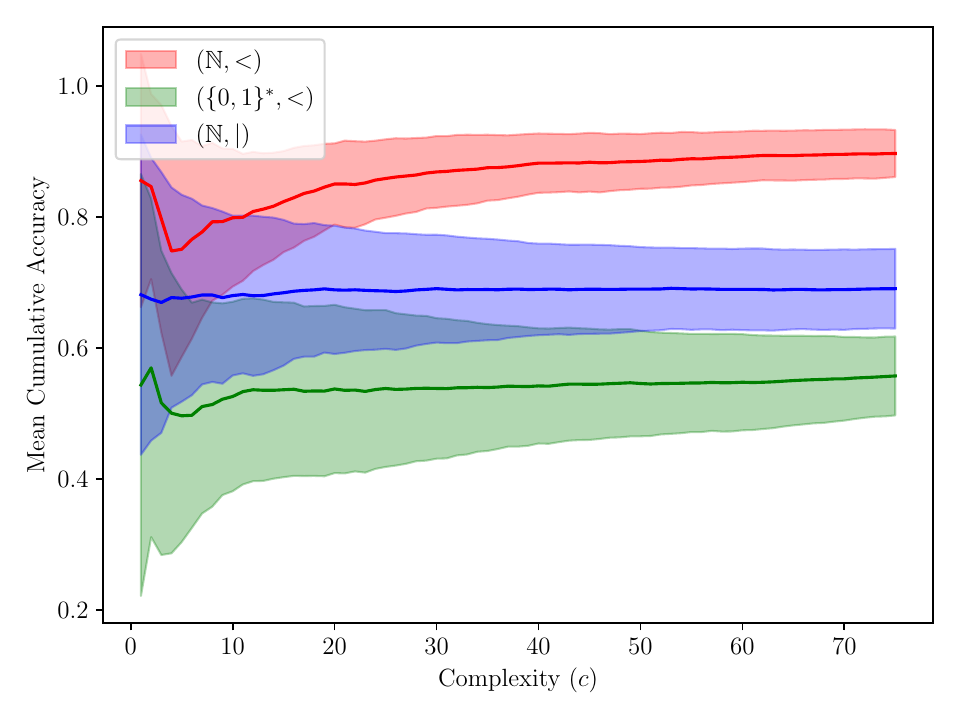} }%
   \end{subfloatrow}\hspace*{\fill}
   \begin{subfloatrow}
      \fcapside[\FBwidth]
        {\caption{}\label{fig:3b}}%
        {\includegraphics[trim={4.5mm 0 0 0mm}, clip, width=0.45\textwidth]{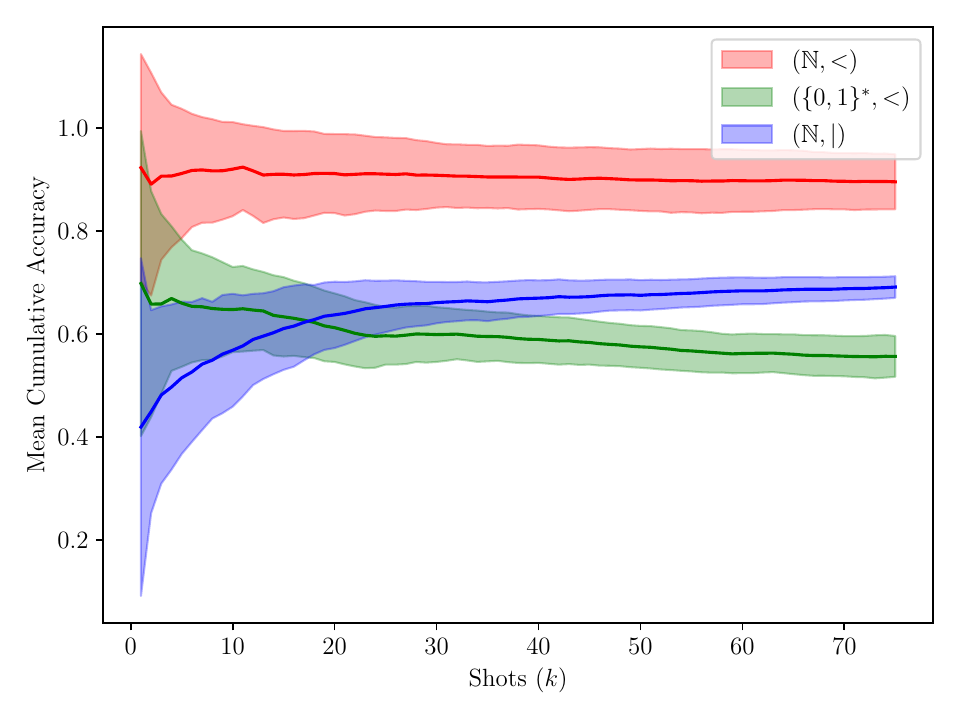}}%
    \end{subfloatrow}\\[-2mm]
   \begin{subfloatrow}
      \fcapside[\FBwidth]
        {\caption{}\label{fig:3c}}%
        { \includegraphics[trim={4.5mm 0 0 0mm}, clip, width=0.45\textwidth]{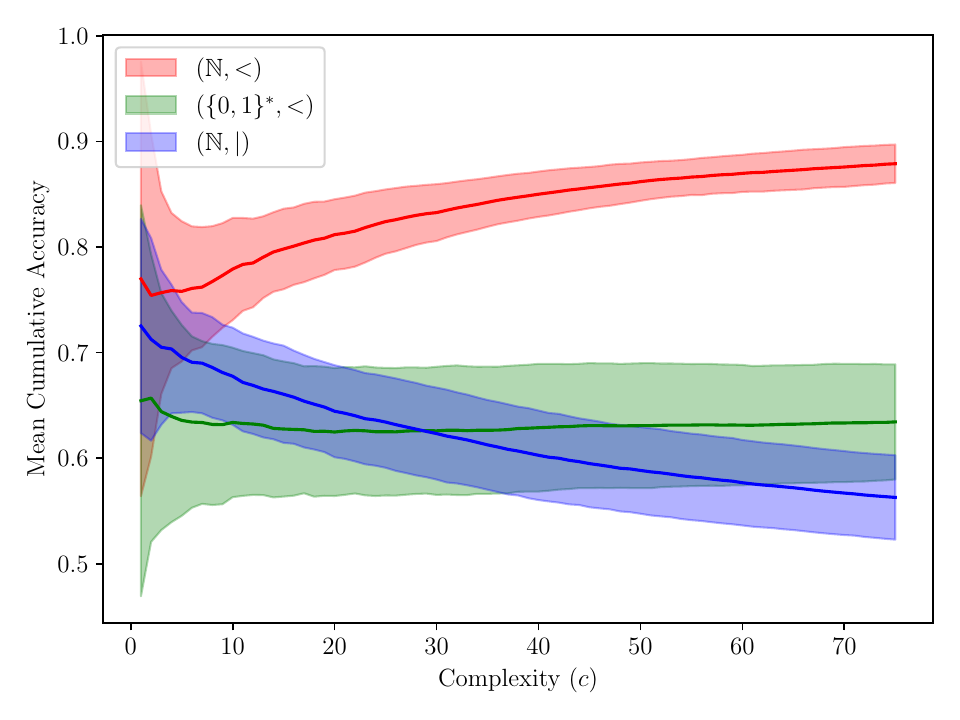}}%
    \end{subfloatrow}\hspace*{\fill}
   \begin{subfloatrow}
      \fcapside[\FBwidth]
        {\caption{}\label{fig:3d}}%
        {\includegraphics[trim={4.5mm 0 0 0mm}, clip, width=0.45\textwidth]{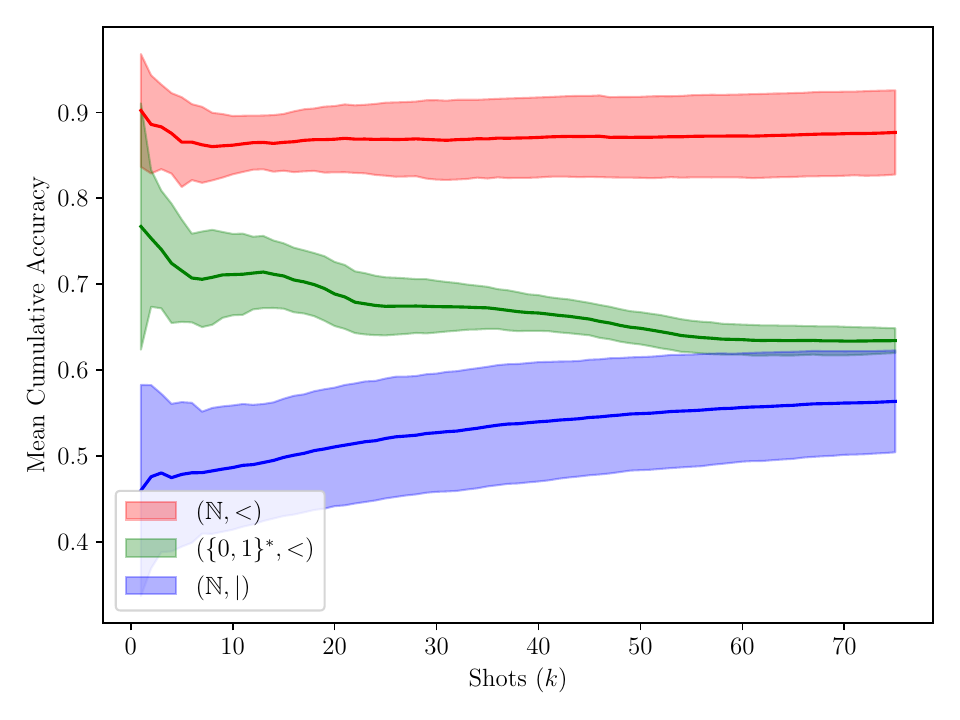}}%
    \end{subfloatrow}\\[-2mm]
    }
  {\caption{Subfigure \ref{fig:3a} and \ref{fig:3c} plots the mean cumulative accuracy (on \textsf{GPT-3.5-Turbo} and \textsf{GPT-4.o-mini} respectively) for the sequence $\set{\L(P_{k,c})}_{k =\set{1, 2, \ldots, 75}}$ against varying complexities for posets \emph{less than} ($(\N, <), (\set{0,1}^*, <)$) and \emph{divides} ($(\N, |)$. Subfigure \ref{fig:3b}  and \ref{fig:3d} similarly describes the same metric for $\set{\L(P_{k,c})}_{c =\set{1, 2, \ldots, 75}}$ against varying shots.\vspace*{-1em}}
        \label{fig:3}}
\end{figure}

    \subsection{Performance on GPT Family}
    \label{sec:gpt}
    We have extended our experiments towards verifying the saturating nature of the evaluation metric by employing LLM such as \textsf{GPT-3.5-Turbo} having context-window 16,385. Instead of performing on the whole dataset, we have constrained the experiments to $75$-shot $75$-complex prompts. Figure \ref{fig:3a} shows that after initial fluctuations ($c\le 30$), performance with prompts from $(\set{0,1}^*,<)$ stabilizes and variation decreases as plot width narrows. Similar trend has also been observed in $(\N, |)$, except the fact that there is no observable fluctuation even in the beginning. However, for the task $(\N,<)$, the model excelled irrespective of the complexity. The abrupt decrease in performance for some initial points resembles $(\set{0,1}^*, <)$. Now keeping the complexity constant on average and increasing the knowledge, for task $(\set{0,1}^*, <)$ (subfigure \ref{fig:3b}), a slightly better performance can be observed compared to subfigure \ref{fig:3a} but it deteriorates, however with decreasing slope. Interestingly, the behavior in $(\N, |)$ shows a sharp increase till $k\le 40$ and stabilizes afterward. Even though $(\N, <)$ has performed significantly well from the beginning it performance did not enhance with increasing knowledge. Because we have conducted our experiment to a compressed dataset, we anticipate a gradual saturation with increasing example.

    \textsf{GPT4.o-mini} with context window of 128k tokens presents an interesting behavior when we consider the experiment on poset $(\N, |)$. Figure \ref{fig:3c} shows mean cumulative accuracy decreases rapidly with increasing complexity till $c=45$ and its descent becomes gradual afterwards {--} showing its incompetency to generalize the representation of arbitrary DAGs. In contrast, Figure \ref{fig:3d} exhibits a slight increase in performance with increasing knowledge. Even though, a perfect saturation could not be observed with compressed data $\set{P_{k,c}}_{k,c =\set{1, 2, \ldots, 75}}$, the subfigures exhibit diminishing descent or gains respectively. On poset $(\set{0,1}^*, <)$, the performance with increasing knowledge (Fig. \ref{fig:3d}) has plateaued after an initial decline.

    For both the \textsf{GPT} models, significant disagreement between the performance of $(\set{0,1}^*, <)$ and $(\N, <)$ (in Fig. \ref{fig:3}) highlights capturing length-dependent lexicography is far-fetched  for such \iffalse heavy\fi models despite their promising performance on somewhat manageable tasks $(\N, <)$.
   
\section{Theoretical Insights}
    \label{sec:5}
    \begin{wrapfigure}[14]{r}{0.45\textwidth}
    \vspace*{-2em}
        \centering
        \includegraphics[width=\linewidth]{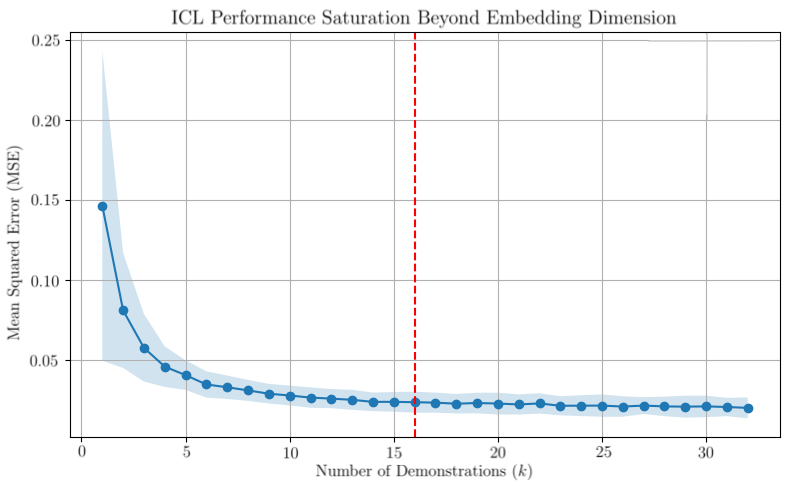}
        \caption{ICL regression performance vs. number of demonstrations $k$. Error bars show $\pm 1$ std across 5 seeds. Vertical dashed line at $k=d=16$ marks the saturation threshold.}
        \label{fig:icl_saturation}
    \end{wrapfigure}    
    Building on the perspective by \citet{dai-etal-2023-gpt}, we frame ICL as a process where pre-trained language models act as meta-optimizers: they generate \textit{meta-gradients} through forward computation over demonstrations and apply these updates via attention mechanisms, effectively simulating parameter tuning without explicit backpropagation. 
    % This duality between attention and optimization provides a principled foundation for analyzing the representational capacity of ICL. 
    % The core mechanism of ICL can be formalized through the lens of linear attention approximation. 
    Consider a pre-trained model $\mathcal{M}$ with embedding dimension $d$, where the attention operation over $k$ demonstration tokens $\{x'_i\}_{i=1}^k$ and their embeddings $\{\mathbf{x}'_i \in \mathbb{R}^d\}$ approximates gradient updates to a base parameter matrix $W_{\text{ZSL}}$. Following \citeauthor{dai-etal-2023-gpt}, the attention output for a query $\mathbf{q}$ can be decomposed into two components:
\[
    \scalemath{0.8}{
    \widetilde{\mathcal{F}}_{\text{ICL}}(\mathbf{q}) = W_{\text{ZSL}}\mathbf{q} + \underbrace{\sum_{i=1}^k \left( W_V\mathbf{x}'_i \otimes W_K\mathbf{x}'_i \right)\mathbf{q}}_{\Delta W_{\text{ICL}}\mathbf{q}},
      }\]
where $W_V, W_K$ are value and key projection matrices. Here, $\Delta W_{\text{ICL}}$ represents the cumulative meta-gradient updates derived from demonstrations. Crucially, each term $(W_V\mathbf{x}'_i) \otimes (W_K\mathbf{x}'_i)$ constitutes a rank-1 matrix, and the rank of $\Delta W_{\text{ICL}}$ is inherently bounded by the number of linearly independent demonstration embeddings. (For more, see Appendix \ref{appendix:ICLMETA}).

\noindent
\textbf{Bounded Representational Capacity of ICL:}
This meta-optimization framework exposes fundamental constraints on ICL’s ability to encode relational structures. As demonstrations increase, the rank of $\Delta W_{\text{ICL}}$ grows linearly until $k$ exceeds $d$, after which additional demonstrations fail to enrich the model’s representational capacity. Formally:  
    \begin{theorem} \label{thm:1}
        For a pre-trained language model $\mathcal{M}$ with parameter $\theta$ and a demonstration context $C=\{(x'_i, y'_i)\}_{i=1}^{k}$ of $k$-shot examples. The update matrix $\Delta W_{ICL}$ in the attention mechanism of ICL (as presented above) has a rank bounded by $\min{(k, d)}$ where $d$ is the embedding dimension. For $k > d$, additional demonstrations do not contribute new information, leading to saturation in the representational capacity of ICL.
    \end{theorem}
\textit{Proof Outline}: Each demonstration contributes a rank-1 update $(W_V\mathbf{x}'_i) \otimes (W_K\mathbf{x}'_i)$. For $k \leq d$, these updates are linearly independent if embeddings $\{\mathbf{x}'_i\}$ span $\mathbb{R}^d$. For $k > d$, embeddings occupy a $d$-dimensional subspace, forcing subsequent updates to lie in the span of prior terms. Consequently, $\Delta W_{\text{ICL}}$ cannot exceed rank $d$, limiting ICL’s capacity to assimilate new information.  (See proof and discussion in Appendix \ref{proof}).

To empirically validate Theorem~\ref{thm:1}, we adapt the regression-based ICL formulation introduced by \citet{guo2024how} and construct a synthetic in-context regression task using a transformer trained on linear functions. The model receives $k$ demonstration pairs $(\mathbf{x}_i, y_i)$ sampled from a fixed linear function, followed by a query input $\mathbf{x}_q$. As shown in Figure~\ref{fig:icl_saturation}, performance improves with increasing $k$ but saturates beyond the embedding dimension $d=16$, confirming our theoretical result that the rank of $\Delta W_{\text{ICL}}$ is upper-bounded by $\min(k, d)$. Full experimental details, including training parameters and additional discussion, are provided in Appendix~\ref{appendix:icl-regression}.

We next also provide a geometric counterpart to the theorem using the concept of \emph{task vectors}. %, where the saturation of task vector clusters in the t-SNE space visually reflects this fundamental limitation.
\section{Task Vector Geometry in Poset ICL}
\label{sec:taskvectors}
To better understand how LLMs represent relational tasks under ICL, we adopt the framework of \emph{task vectors} introduced by \citet{hendel2023context}, which characterizes ICL as a two-step process: compressing a prompt into a fixed-size \textit{task vector} $\theta$, and applying a query-conditioned function $f(x; \theta)$ to generate predictions.

Formally, given a $k$-shot, $c$-complex prompt $P_{k,c}$, we first adapt it to $\widetilde{P}_{k,c}$ such that $\widetilde{P}_{k,c}$ does not contain the instruction $I$ (Fig. \ref{fig:gram1}) explicitly. In turn, the minimal set of demonstrations has now been labeled. We define the task vector $\theta(\widetilde{P}_{k,c}) \in \mathbb{R}^d$ as the hidden representation of a sentinel token (e.g., ``$\rightarrow$'') at a fixed transformer layer $\ell$:
$
\theta(\widetilde{P}_{k,c}) := \mathrm{Enc}_\ell(\widetilde{P}_{k,c})[t_{\rightarrow}],
$
where $\mathrm{Enc}_\ell(\cdot)$ denotes the model's (e.g. \textsf{Llama3}) embedding at layer $\ell$, and $t_{\rightarrow}$ is the index of the delimiter token (See Appendix~\ref{appendix:task_vectors} for detail).

\begin{figure}[!ht]
    % \vspace*{-3mm}
  \ffigbox{%
    \begin{subfloatrow}
      \fcapside[\FBwidth]
        {\caption{}\label{fig:6a}}%
        {\includegraphics[trim={2mm 0 0 0}, clip, width=0.30\textwidth,height=0.3\textwidth]{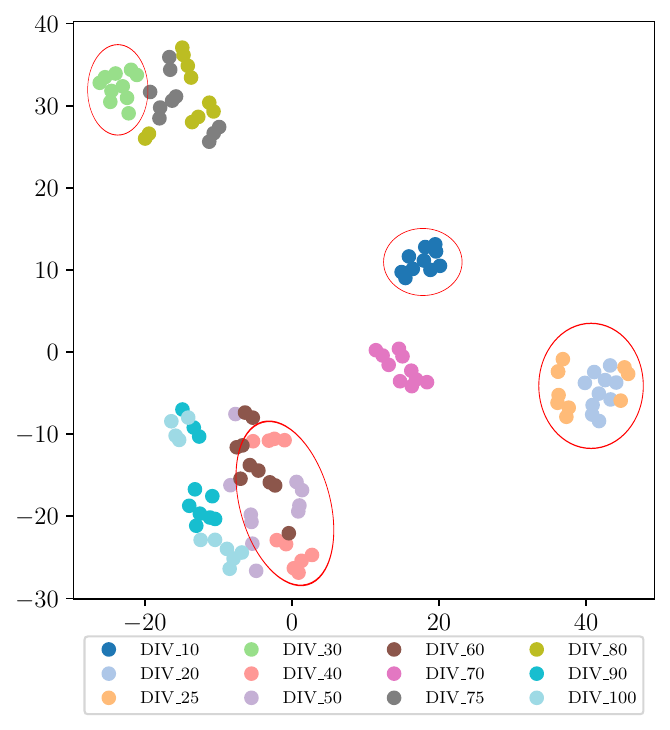}}%
   \end{subfloatrow}\hspace*{\fill}
    \begin{subfloatrow}
      \fcapside[\FBwidth]
        {\caption{}\label{fig:6b}}%
        {\includegraphics[width=0.30\textwidth,height=0.30\textwidth]{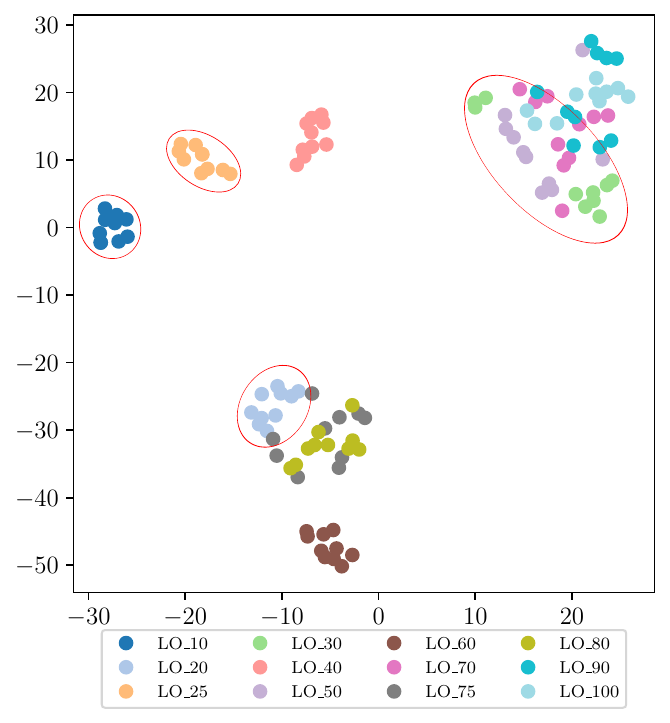}}%
   \end{subfloatrow}\hspace*{\fill}
    \begin{subfloatrow}
      \fcapside[\FBwidth]
        {\caption{}\label{fig:6c}}%
        {\includegraphics[trim={2mm 0 0 0}, clip, width=0.30\textwidth,height=0.30\textwidth]{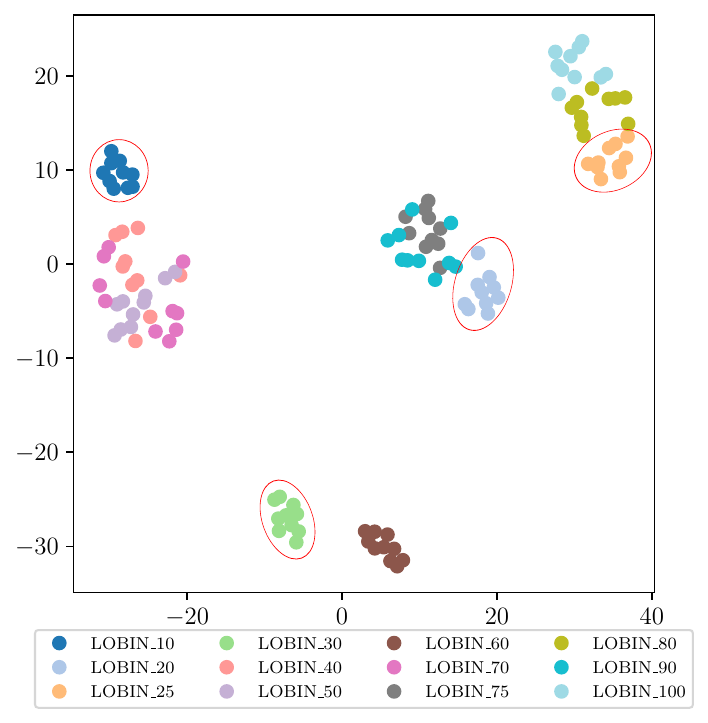}}%
   \end{subfloatrow}\\[-1mm]
    \begin{subfloatrow}
      \fcapside[\FBwidth]
        {\caption{}\label{fig:6d}}%
        {\includegraphics[width=0.30\textwidth,height=0.30\textwidth]{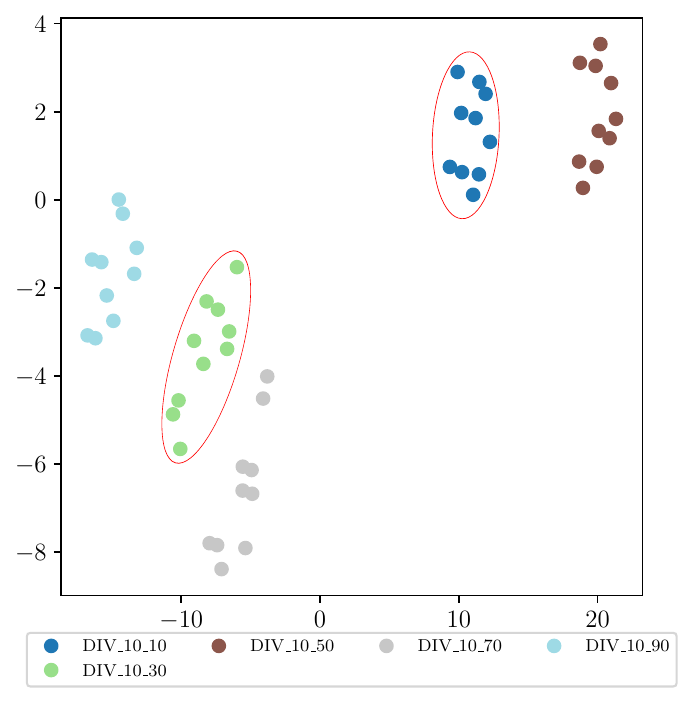}}%
   \end{subfloatrow}
    \begin{subfloatrow}
      \fcapside[\FBwidth]
        {\caption{}\label{fig:6e}}%
        {\includegraphics[trim={2mm 0 0 0}, clip, width=0.30\textwidth,height=0.30\textwidth]{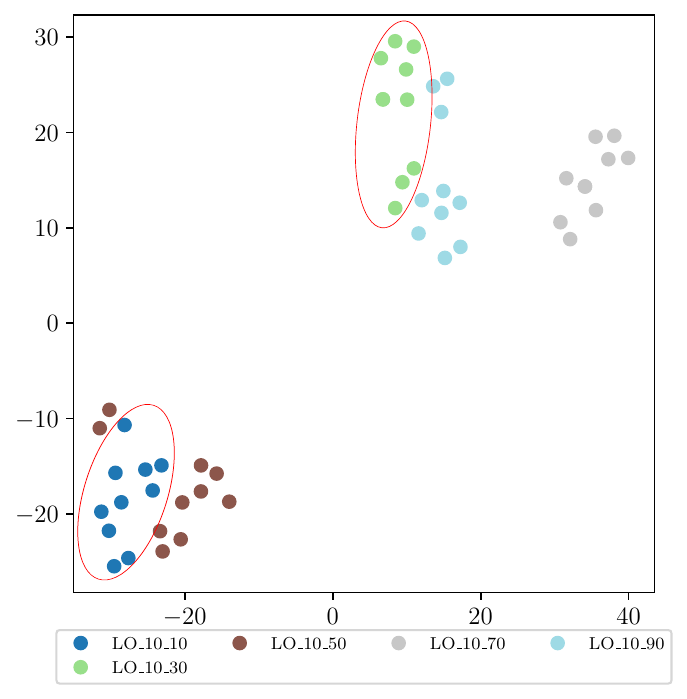}}%
   \end{subfloatrow}
   \begin{subfloatrow}
      \fcapside[\FBwidth]
        {\caption{}\label{fig:6f}}%
        {\includegraphics[trim={2mm 0 2mm 0mm}, clip, width=0.30\textwidth,height=0.30\textwidth]{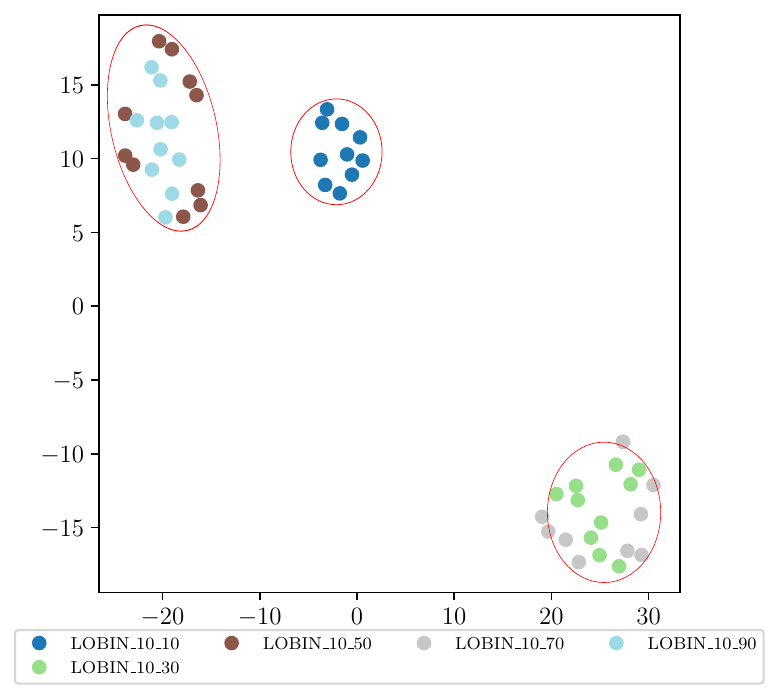}}%
    \end{subfloatrow}
    }
  {\caption{t-SNE projections of task vectors for each task: \texttt{DIV}, \texttt{LO}, \texttt{LOBIN}. (Subfigures \ref{fig:6a}–\ref{fig:6c}): fixed complexity, varying shots $k \in [10, 100]$. (Subfigures \ref{fig:6d}–\ref{fig:6f}): fixed $k = 10$, varying complexity $c \in [10, 90]$. Circled clusters denote lower-shot or lower-complexity conditions. Overlap at higher values reflects latent representational saturation. (Based viewed in 300\% zoom.)\vspace*{-1em}}
        \label{fig:tsne-shots}}
\end{figure}

We conduct geometric analysis of task vectors under two settings: (i) \emph{Varying demonstrations with fixed complexity}: For each $k \in \{10, 20, 25, \ldots, 100\}$, we construct 10 prompts using the same demonstration set and 10 distinct queries, producing 10 task vectors per case. These are projected via t-SNE and plotted for each task—\texttt{DIV} ($(\N, \mid)$), \texttt{LO} ($(\N, <)$), and \texttt{LOBIN} ($(\{0,1\}^*, <)$)—using the format \texttt{Task\_k} (e.g., \texttt{LO\_50}). 
(ii) \emph{Varying complexity with fixed demonstrations}: For each complexity level $c \in \{10, 30, 50, 70, 90\}$, we fix $k=10$ and generate 10 prompts by combining the same demonstrations with queries of the specified complexity. These are similarly projected via t-SNE, and labeled as \texttt{Task\_10\_c} (e.g., \texttt{DIV\_10\_30}) indicates the task, fixed demonstrations, and complexity level. In both settings, low-shot or low-complexity cases are highlighted using circles.

As shown in Figure~\ref{fig:6a}-\ref{fig:6c}, all three tasks exhibit a characteristic pattern of geometric convergence. In the low-shot regime ($ k \leq 30 $, circled), task vector clusters are relatively well-separated, indicating distinct representations when contextual support is limited. However, as $ k $ increases, the corresponding clusters begin to collapse and increasingly overlap with those from lower-shot settings. This drift toward a shared latent region suggests diminishing representational changes as the number of demonstrations grows, consistent with saturation effects. \emph{Task-specific} behaviors are also evident. For the \texttt{DIV} task, which involves a non-trivial DAG structure, the representations gradually merge into a contiguous manifold beyond $ k=60 $. In contrast, the \texttt{LO} task shows early overlap from $ k=40 $ onward, reflecting the simpler, transitive nature of total orders. The \texttt{LOBIN} task exhibits delayed convergence, with binary encodings preserving more structural variation at higher $ k $ values.

% These patterns visually substantiate Theorem~\ref{thm:1} that ICL’s meta-gradient updates saturate beyond the embedding dimension $d$, as reflected by the overlap of task vector clusters in t-SNE space.
The overlap of task vector clusters in t-SNE space visually substantiate Theorem~\ref{thm:1} that ICL’s meta-gradient updates saturate beyond the embedding dimension $d$.

% , which establishes that ICL's meta-gradient updates are rank-limited by the embedding dimension $d$, implying that beyond this threshold, additional demonstrations yield no new representational capacity. The saturation of task vector clusters in t-SNE space visually reflects this fundamental limitation.

Figures \ref{fig:6d}, \ref{fig:6e} and \ref{fig:6f} show similar patterns when varying complexity. Across all tasks we observe that task vectors from lower-complexity prompts ($ c = 10, 30 $) tend to form distinct clusters. However, as complexity increases, the representations become progressively less distinguishable and begin to overlap with those of lower complexity. This convergence suggests that higher complexity levels do not always induce richer or structurally unique task encodings within the model’s latent space. \emph{Task-specific} behaviors again emerge. For the \texttt{DIV} task, which involves arbitrary DAGs, the latent representations at higher complexities ($ c \geq 70 $) collapse noticeably, indicating a possible limit in the model’s ability to understand and represent complex relationships that involve multiple reasoning steps. The \texttt{LO} task saturates earlier, with convergence emerging from $c=50$, likely due to limited structural variation in total orders, whereas \texttt{LOBIN} maintains moderate separation, reflecting the richer combinatorics of binary encodings. We defer this study on \textsf{Pythia-2.8B} \citet{biderman2023pythia} and \textsf{Llama3} (along with incorporating instruction $I$ to $\widetilde{P}_{k,c}$) in Appendix \ref{appendix:tSNEwithRelDesc}.
% The \texttt{L} task exhibits earlier convergence, with representations saturating from $ c=50 $ onward, likely reflecting the limited incremental variation possible within total orders. Interestingly, the \texttt{LOBIN} task retains moderate separation across complexity levels, potentially due to the nuanced combinatorial structure introduced by binary encodings.

\section{Conclusion and Future Directions}
\label{sec:6}
    We presented the first study of ICL in LLMs for posets, addressing a key gap in their reasoning beyond functions. Our research introduced a novel evaluation framework using $k$-shot $c$-complex prompts to assess how LLMs infer hierarchical structures, such as the less than $(\mathbb{N}, <)$ and divisibility $(\mathbb{N}, |)$ posets, which are fundamental in mathematics and real-world applications (e.g., genealogical trees, set inclusion). These posets were chosen for their distinct properties—path vs arbitrary DAG structures—providing a rigorous testbed for evaluating transitive and anti-symmetric reasoning. The exclusion of exponential-growth posets like $(2^S, \subseteq)$ was necessitated by computational constraints, but the methodology remains extensible to such cases.  
    Our study reveals fundamental limitations of ICL for poset reasoning: while LLMs learn basic linear orders (e.g., $(\mathbb{N}, <)$ from few-shot examples, performance plateaus with increasing complexity. For DAG-structured posets like $(\mathbb{N}, |)$, ICL shows greater limitations, failing to generalize beyond initial demonstrations. Even large-context models (e.g., GPT-4) struggle with lexicographic orders $(\{0,1\}^*, <)$, exposing ICL's inherent constraints in learning hierarchical relations without parameter updates.  
    Theoretical analysis tied this limitation to the bounded rank of meta-gradient updates, highlighting a fundamental constraint on ICL’s representational capacity. Task vector geometry further confirmed this saturation, showing collapsed latent representations for high-complexity prompts. Future work should explore hybrid models that combine neural and formal reasoning. By bridging theoretical foundations with empirical rigor, this study advances the discourse on LLM's limitations and opportunities in relational learning.
% \clearpage
\section{Limitation}
Our research explores ICL on various language models for partially ordered relations. However, it is essential to acknowledge certain limitations. As we have utilized quantized version (\texttt{Q4\_0}) of these models due to limitations in computational resources.  This research also could delve into investigation of performance on simpler architectures like vanilla Transformer or LSTM networks, as emphasized by \citet{Satwick2024} which may provide insights and verification towards our claim from a more foundational level.

While this research indicates that existing mathematical models for scattered linear orders \citep{LoAut1, LoAut2} are not directly implementable via ICL, further analysis could yield insights into ICL's computational capabilities.

Readers might be concerned about our experiments using GPT to compress the evaluation dataset to one-fourth of its original size. However, our empirical framework for assessing ICL demonstrates justifiable analogous behavior in performance, similar to what is observed in local LLMs, even within this constrained scenario with limited data, a measure towards saving expenses in the subscription-based GPT models.

\bibliography{main}
\bibliographystyle{tmlr}

\clearpage
\appendix
\section*{Appendix}
\section{Results on Fine-tuning and Illustrative Prompts}
\label{ssec:A1}
\begin{figure}[H]
    \vspace*{-3mm}
    \ffigbox{
    \begin{subfloatrow}
        \fcapside[\FBwidth]
        {\caption{}\label{fig:finetunea}}%
        {\includegraphics[trim={4.5mm 0 0 0mm}, clip, width=0.45\textwidth]{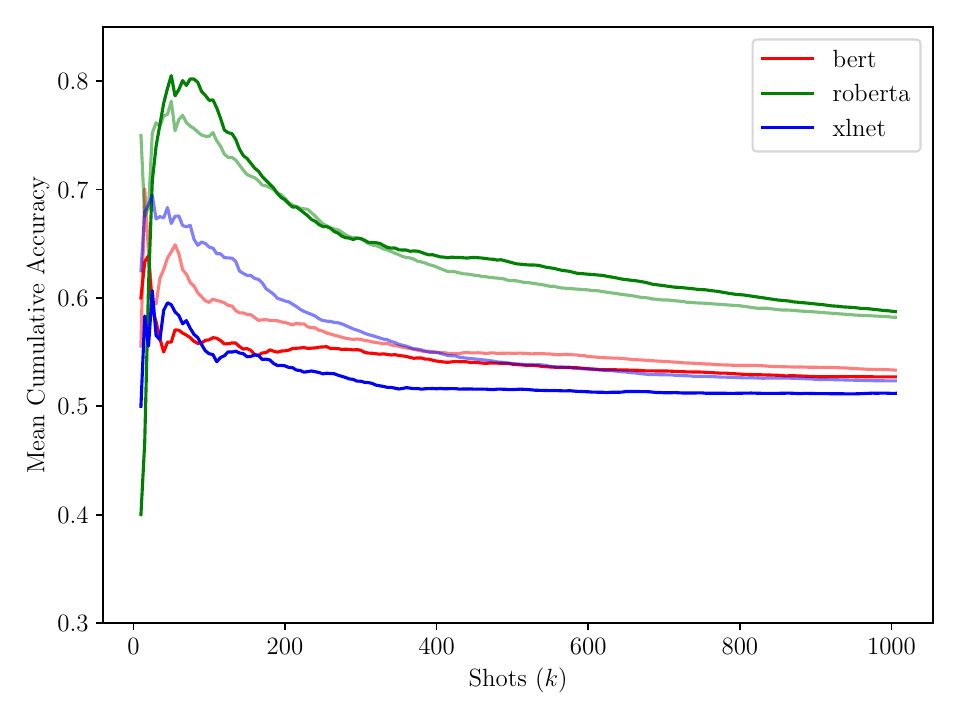}}%
    \end{subfloatrow}
    \begin{subfloatrow}\hspace*{\fill}
      \fcapside[\FBwidth]
       {\caption{}\label{fig:finetuneb}}%
       {\includegraphics[trim={4.5mm 0 0 0mm}, clip, width=0.45\textwidth]{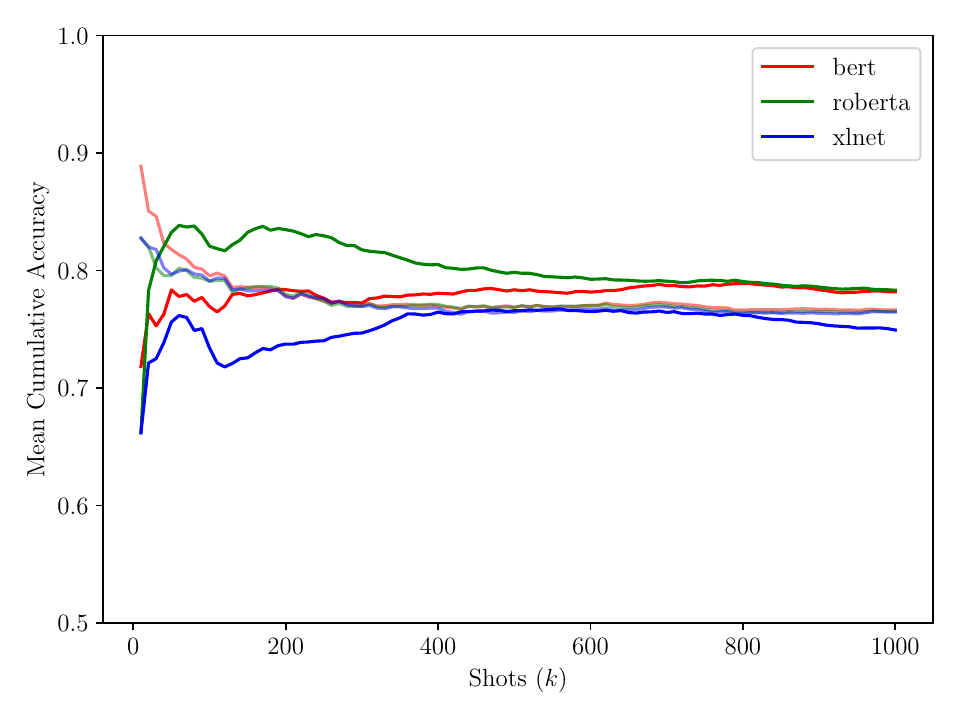}}%
    \end{subfloatrow}
    }
    {\caption{Subfigure \ref{fig:finetunea} represents the mean cumulative accuracy plot while fine-tuned for linear order $(\N, <)$. And \ref{fig:finetuneb} represents the same metric for the partial order $(\N, \mid)$.}
    \label{fig:finetune}}
    \end{figure}

    Large (neural) language models can generally be classified into two categories: those with encoder-only architectures and those with decoder-only architectures. Due to the ability to produce self-attentive encoded representations, the former is widely adopted for language classification tasks. In contrast, the latter, being auto-regressive in nature, is primarily used for language generation. The task description, given by $D$, in Figure \ref{fig:gram1} may drive us to analyse ICL on encoder-only LLMs and do fine-tuning for the other category. However, following the opposite approach does not result much improvements which can be implied from the plots of finetuned LLMs as presented in Figure \ref{fig:finetune}.
    In the fine-tuning experiments, because of significant training time, we have performed similar experiments by keeping the complexity constant concerning the provided knowledge. Consider a sequence of $k$-shot $k$-complex prompts $\set{P_{k,k}}_{k=\set{10,15,\ldots,1000}}$. The fine-tuning involves producing the examples of the prompt $P_{k,k}$ as training samples alike binary classification problem while inference was based on the range $(\set{1, 2, 3,\dots, 2k}, <)$. Figure \ref{fig:finetunea} shows a behavior similar to the earlier experiment (Fig. \ref{fig:1}) for all the encoder-based models. The plots with faded lines show the behavior of test cases drawn from the training range but not included in training samples. An observable saturation can be noted for the models BERT and XLNet after $k=600$, while RoBERTa shows a decaying performance which becomes milder as $k$ increases beyond $800$.\\
    As we continue for the poset $(\mathbb{N}, \mid)$ under similar setup, Figure \ref{fig:finetuneb}, presents mean cumulative accuracy (y-axis) over training range (x-axis). The plot for BERT, RoBERTa, and XLNet, with solid lines (and faded lines) represents average training performance (and variations on test cases). RoBERTa consistently outperforms BERT and XLNet throughout training, stabilizing around the training range $400$, while BERT and XLNet display similar trends, with XLNet achieving lower accuracy overall. This pattern suggests that RoBERTa adapts more efficiently during fine-tuning, although all three models eventually reach a performance plateau where further fine-tuning yields minimal improvement. Despite initial benefits of fine-tuning, performance saturation indicates, these models fully exploit contextual knowledge available in training data early on.
    
    Table \ref{tab:Models} specifies the open-sourced language models that we have used during our ICL experiment which are available in \textsf{ollama} library. The experiment is expensive in terms of both resource and computation time. The experiment has been conducted on NVIDIA GeForce RTX 3090 series.We use BERT and RoBERTa as key encoder-based models for NLP, leveraging bidirectional context, unlike autoregressive LLMs. This makes them well-suited for fine-tuning in binary classification tasks like ours. For completeness, we also fine-tune XLNet, which captures bidirectional context using an autoregressive approach. Fine-tuning is conducted for up to 100 epochs with Early Stopping. See table \ref{tab:Models} for details.
    \begin{table*}
        \begin{tabular}{llcc}
        \hline
             Models & Parameters & Avg. Inference Time & Avg. Fine-tune Time\\\hline
             Gemma2 \citet{Gemma2Il} & 9.24B & 9.87-13.03 sec & \texttimes\\ %div:9.87, lo: 12.85, bin:13.03
             Lamma3  \citet{llama} & 8.03B & 5.32-7.9 sec & \texttimes\\ %div:5.32, lo:6.23, bin: 7.9
             Mathstral \citet{mathstral} & 7.25B & 4.98-15.23 sec & \texttimes\\ %div:4.98, lo: 6.87, bin: 15.23
             Qwen2-math \citet{qwen25} & 7.62B & 3.08-30.88 sec & \texttimes\\ %lo:8.33, div:3.08
             Phi3 \citet{phi3} & 14B & 1.08-7.05 sec & \texttimes\\ %lo: 1.08, div: 4.52
             GPT3.5 Turbo & \texttimes & 3.32-5.80 sec & \texttimes\\ %(LO: 4.55, 5.80, div: 3.32)
             GPT4.o Mini & \texttimes & 5.45-10.14 sec & \texttimes\\ \hdashline %(LO:6.90, 10.14, div: 5.54)
             BERT \citet{devlin-etal-2019-bert} & 110M & \texttimes & 8.28 - 9.37 min \\
             RoBERTa \citet{liu2019roberta} & 125M & \texttimes & 10.42 - 13.01 min \\
             XLNet \citet{xlnet} & 110M & \texttimes & 7.36 - 8.43 min \\\hline
         % \hline
        \end{tabular}
        \caption{Model specifications and inference/fine-tuning times are presented in this table. The reported time represents the average duration required to process a single prompt, either for in-context learning (ICL) or fine-tuning using examples from a single prompt for respective models.}
        \label{tab:Models}
    \end{table*}
    % On the other hand, we have used \textsf{BERT}, \textsf{RoBERTa} and BERT-inspired autoregressive \textsf{XLNet} for fine-tuning purpose which took 
    
    Analogous to the Hasse diagram provided in Figure \ref{fig_Hasse}, the below block illustrates a $4$-shot $2$-complex prompt $P_{4,2}$ on $(\N, \mid)$. Please note, even though we considered the cardinality of $P_{k,c}[T^\prime]$ are $50$ and $30$ for $(\N, <)$ and $(\N, \mid)$ respectively in Section \ref{sec:4}, for trivial cases where it was not possible to sample the required number of test cases, we generated as many distinct cases as feasible. Example 2 and 3, similarly, demonstrate the prompts for linear order of $P_{26,101}$ in the $(\N, <)$ and $P_{33,84}$ in the $(\set{0,1}^*, <)$.\\

        \begin{tcbraster}[raster columns=2,raster valign=top]
      \begin{tcolorbox}[colback=white,colframe=blue!5,title=\textcolor{black}{Example Prompt 1}, fontupper=\fontsize{8.5pt}{5pt}\selectfont, fontlower=\fontsize{8.5pt}{5pt}\selectfont,left=-1pt, right=-1pt]
      \label{example1}
        \texttt{\hspace*{-2mm}
          Relation Description: There is a relation divides `$\mid$' between integers composed of digits "0", "1", "2", "3", "4", "5", "6", "7", "8" and "9".\\
          Given such integers x, y, and z, the relation has the following properties:\\
          \hspace{1cm}(a) if x$\mid$y is true, then y$\mid$x is false,\\
          \hspace{1cm}(b) if both x$\mid$y and y$\mid$z are true, then x$\mid$z is true, and\\
          \hspace{1cm}(c) x$\mid$x is always true, for any x.\\
          Examples: "1 $\mid$ 1", "1 $\mid$ 2", "2 $\mid$ 2", "1 $\mid$ 3", "3 $\mid$ 3", "2 $\mid$ 4", "4 $\mid$ 4"\\[1.2mm]
          Task Description: Given the above information, determine if \{integer 1\}\ $\mid$ \ \{integer 2\} is true.\\
          \hspace{1cm} If it is true, your answer must be "\{integer 1\} \ $\mid$ \  \{integer 2\}: true".\\
          \hspace{1cm} If it is false, your answer must be "\{integer 1\} \ $\mid$ \ \{integer 2\}: false".\\
          \hspace{1cm}If you do not know if it is true or false, you answer must be "\{integer 1\}\ $\mid$ \ \{integer 2\} : unknown".\\[1mm]  
          Follow the specified format for answering:\\
          1 $\mid$ 5\\
          2 $\mid$ 6\\
          5 $\mid$ 6\ \ \dots
          }
          \tcblower
          Response by \textsf{Gemma2}:\\
          \texttt{
          "1 $\mid$ 5 : true\\
          2 $\mid$ 6 : true\\
          5 $\mid$ 6 : \textcolor{gray}{true}\ \ \dots"
          }
      \end{tcolorbox}
    \begin{tcolorbox}[colback=white,colframe=blue!5,title=\textcolor{black}{Example Prompt 2}, fontupper=\fontsize{8.5pt}{5pt}\selectfont, fontlower=\fontsize{8.5pt}{5pt}\selectfont,left=-1pt, right=-1pt]
      \label{example2}
        \texttt{\hspace*{-2mm}
          Relation Description: There is a relation less than "<" between strings made of decimal digits "0", "1", "2", "3", "4","5", "6", "7", "8" and "9".\\
          Given such integers x, y, and z, the relation has the following properties:\\
          \hspace{1cm}(a) if x<y is true, then y<x is false,\\
          \hspace{1cm}(b) if both x<y and y<z are true, then x<z is true, and\\
          \hspace{1cm}(c) x<x is always false, for any x.\\[1.2mm]
          Examples: "0 < 1", "1 < 2", "2 < 3", \dots, "24 < 25", "25 < 26"\\[1.2mm]
          Task Description: Given the above information, determine if \{integer 1\}\ < \ \{integer 2\} is true.\\
          \hspace{1cm} If it is true, your answer must be "\{integer 1\} \ < \  \{integer 2\}: true".\\
          \hspace{1cm} If it is false, your answer must be "\{integer 1\} \ < \ \{integer 2\}: false".\\
          \hspace{1cm}If you do not know if it is true or false, you answer must be "\{integer 1\}\ < \ \{integer 2\} : unknown".\\[1mm]  
          Follow the specified format for answering:\\
          121 < 16\\
          22 < 85\\
          73 < 15\\
          122 < 18\ \ \dots
          }
          \tcblower
          Response by \textsf{Llama3}:\\
          \texttt{
          "121 < 16 : \textcolor{gray}{true}\\
          22 < 85 : true\\
          73 < 15 : false\\
          122 < 18 : false\ \ \dots"
          }
          \\[2mm]\textcolor{gray}{\rule{\textwidth}{.1pt}}\\[2mm]
          Response by \textsf{Phi3}:\\
          \texttt{
          "The numbers that are less than the first number but greater than or equal to it is called 'lesser'. So, if x=2 and y = 3, then the range of $x^4 - 7x^3 + 19x^2 + 60$ is $[2x-5, 8]$ because when you substitute in $x=-5$ ,you get: $p(x) = a (x-5)^4 - 7a(x+5)-19a(x+5)+60$.\dots% If we compare this with the given inequality, then we have $a(x-5)^4 - 7a(x+5)-19a(x+5)+60 < 20$. Since $(x-5)^4$ is always positive and greater than or equal to zero for any $x$, when it's raised to the power of four, it will be a very large number if $x$ is negative.\\
          % Hence, we have:\\
          % $7(a)(x^3 - 19ax^2 + 60) < 20$
          "
          }
      \end{tcolorbox}
    \end{tcbraster}
        \begin{tcolorbox}[colback=white,colframe=blue!5,title=\textcolor{black}{Example Prompt 3}, fontupper=\fontsize{8.5pt}{5pt}\selectfont, fontlower=\fontsize{8.5pt}{5pt}\selectfont,left=-1pt, right=-1pt]
          \label{example3}
            \texttt{\hspace*{-2mm}
              Relation Description: There is a relation less than "<" between binary strings made of characters "0" and "1".\\
              Given such strings x, y, and z, the relation has the following properties:\\
              \hspace{1cm}(a) if x<y is true, then y<x is false,\\
              \hspace{1cm}(b) if both x<y and y<z are true, then x<z is true, and\\
              \hspace{1cm}(c) x<x is always false, for any x.\\[1.2mm]
              Examples: "0 < 1", "1 < 10", "10 < 11", \dots, "11111<100000", "100000<100001"\\[1.2mm]
              Task Description: Given the above information, determine if \{string 1\}\ < \ \{string 2\} is true.\\
              \hspace{1cm} If it is true, your answer must be "\{string 1\} \ < \  \{string 2\}: true".\\
              \hspace{1cm} If it is false, your answer must be "\{string 1\} \ < \ \{string 2\}: false".\\
              \hspace{1cm}If you do not know if it is true or false, you answer must be "\{string 1\}\ < \ \{string 2\} : unknown".\\[1mm]  
              Follow the specified format for answering:\\
              "1010001 < 111111"\\
              "100010 < 1000011"\\
              "1001011 < 110010"\ \  \dots
              }
              \tcblower
              Response by \textsf{Qwen2-math}:\\
              \texttt{
              "1010001 < 111111: false\\
              100010 < 1000011: \textcolor{gray}{false}\\
              1001011 < 110010: false\ \ \dots"
              }
          \end{tcolorbox}

\section{ICL as Meta-Optimizer}
\label{appendix:ICLMETA}
Let $W, \Delta W \in \mathbb{R}^{d_{out}\times d_{in}}$ denotes the parameter weights and update matrix of a vanilla feed-forward linear network. On receiving an input representation $\vec{x}\in \mathbb{R}^{d_{in}}$ a single step of gradient descent can be expressed as \[\mathcal{F}(\vec{x}) = \left(W + \Delta W\right)\vec{x}.\] Expanding $\Delta W$ can be unfolded using $\sum_i \vec{e}_i\otimes \vec{x}'_i$ where $\vec{e}_i$ is the negative of the gradient multiplied by an learning rate and $\vec{x}'_i$ be the prior input representation. This allows us to rewrite $\mathcal{F}(\vec{x}) = W\vec{x}+\sum_i( \vec{e}_i\otimes \vec{x}'_i)\vec{x} = W\vec{x} + \sum_i \vec{e}_i \left({\vec{x}^\prime_i}^\top\vec{x}\right) = \operatorname{LinAttn}(E, X', \vec{q})$, where $\operatorname{LinAttn}(V,K, \vec{q})$ denotes linear attention.\\
    To facilitate a clear qualitative analysis of ICL, softmax attention is often approximated with linear attention. This relaxation has been instrumental in prior work investigating the representational capabilities of ICL. During the course of this discussion, we will follow the assessment provided by \citeauthor{dai-etal-2023-gpt}. Given $W_Q, W_K$ and $W_V \in \mathbb{R}^{d\times d}$, being the projection matrices for computing the (attention) queries, keys, and values, respectively, where $d$ denotes the embedding dimension; let $X$ denotes the input representations of query tokens occurring prior to the current query token $t$. Suppose that the token $t$ has input representation $\vec{x}$ and been represented by the query vector $\vec{q} = W_Q\vec{x}$. Let $X^\prime$ denotes the input representations of the example tokens in a ICL prompt, then avoiding the dimension scaling a softmax attention can be expressed as 
    \[\mathcal{F}_{ICL}(\vec{q}) = W_V[X^\prime;X]\operatorname{softmax}\left(W_K[X^\prime;X]^\top\vec{q}\right).\]
    Replacing the softmax attention with a linear attention gives rise to 
    \begin{align*}\small
        \mathcal{F}_{ICL}(\vec{q}) &\approx W_V[X^\prime;X]\left(W_K[X^\prime;X]^\top\vec{q}\right)\\
        &= W_VX(W_KX)^\top\vec{q} + W_VX'(W_KX')^\top\vec{q}\\
        &= W_{ZSL}\vec{q}+\operatorname{LinAttn}(W_KX^\prime,W_VX^\prime, \vec{q})\\
        &= \overset{\sim}{\mathcal{F}}_{ICL}(\vec{q})
    \end{align*} where, $W_{ZSL}\vec{q}=W_VX(W_KX)^\top\vec{q}$. This formulation helps to rewrite $\overset{\sim}{\mathcal{F}}_{ICL}(\vec{q})$ as follows representing attention to examples tokens is equivalent to parameter update an amount of $\Delta W_{ICL}$ affecting $W_{ZSL}$:
\begin{align*}\small
    \overset{\sim}{\mathcal{F}}_{ICL}(\vec{q}) &= W_{ZSL}\vec{q} + \operatorname{LinAttn}(W_KX^\prime,W_VX^\prime, \vec{q})\\
    &= W_{ZSL}\vec{q} + \sum_i \left((W_V\vec{x'}_i) \otimes (W_K\vec{x'}_i)\right)\vec{q}\\
    &= W_{ZSL}\vec{q} + \Delta W_{ICL}\vec{q}.
\end{align*}
    
    \noindent
    Based on this observation, we present the proof of Theorem \ref{thm:1} below. % bounded approximated capacity of ICL (see proof in \ref{proof}) supporting its saturating performance in Theorem 1.

\subsection{Proof of Theorem \ref{thm:1}}
\label{proof}
    \begin{proof}
    ICL computes the update matrix $ \Delta W_{\text{ICL}} $ in the attention mechanism as:
    \[
    \Delta W_{\text{ICL}} = \sum_{i=1}^k (W_V \vec{x}_i') \otimes (W_K \vec{x}_i')^\top,
    \]
    where, $ W_V \in \mathbb{R}^{d^{`} \times d} $ is the value projection matrix, $ W_K \in \mathbb{R}^{d^{`} \times d} $ is the key projection matrix,  $ \vec{x}_i' \in \mathbb{R}^d $ represents the embedding of the $ i $-th demonstration token.
    
    The term $ (W_V \vec{x}_i') \otimes (W_K \vec{x}_i')^\top $ is an outer product, which contributes a rank-1 matrix to $ \Delta W_{\text{ICL}} $. Thus, the rank of $ \Delta W_{\text{ICL}} $ is at most the number of linearly independent embeddings $ \vec{x}_i' $. Formally, the rank is bounded as:
    \[
    \text{rank}(\Delta W_{\text{ICL}}) \leq \min(k, d).
    \]
    
    For $ k \leq d $, each demonstration contributes a unique, linearly independent term, and the rank of $ \Delta W_{\text{ICL}} $ increases linearly with $ k $. However, for $ k > d $, the token embeddings $ \vec{x}_i' $ lie in a $ d $-dimensional space, and any additional embedding $ \vec{x}_{j}' $ can be expressed as a linear combination of the first $ d $ embeddings:
    \[
    \vec{x}_j' = \sum_{i=1}^d \alpha_i \vec{x}_i', \quad \text{for } j > d.
    \]
    Substituting this into $ (W_V \vec{x}_j') \otimes (W_K \vec{x}_j')^\top$, we observe that the corresponding matrix lies in the span of the previous $ d $ terms:
    {\small\begin{align*}
    (W_V \vec{x}_j') \otimes (W_K \vec{x}_j')^T \in \text{span}\{(W_V \vec{x}_i') \otimes (W_K \vec{x}_i')^\top \mid i \leq d\}.
    \end{align*}}
    Thus, for $ k > d $, the rank of $ \Delta W_{\text{ICL}} $ saturates at $ d $, and the representation of the context does not improve with additional tokens.
    
    Alternatively can also illustrates the saturation as:
    Let $ k = d + r $, where $ r > 0 $. The additional terms $ \vec{x}_{d+1}', \vec{x}_{d+2}', \ldots, \vec{x}_{d+r}' $ contribute the following:
    \begin{align*}
        \Delta W_{\text{ICL}} =& \sum_{i=1}^d (W_V \vec{x}_i') \otimes (W_K \vec{x}_i')^\top\\
        &\qquad + \sum_{j=d+1}^{d+r} (W_V \vec{x}_j') \otimes (W_K \vec{x}_j')^\top.
    \end{align*}
    However, since $ \text{rank}(\Delta W_{\text{ICL}}) \leq d $, the second summation does not increase the rank of $ \Delta W_{\text{ICL}} $ and introduces redundancy.
    \end{proof}
    \begin{remark}
    This theorem aligns with our empirical observations of performance saturation. \iffalse in poset reasoning tasks.\fi As complexity increases (e.g., longer chains in Hasse diagrams), the model’s ability to infer transitive or antisymmetric relations depends on its capacity to encode linearly independent updates. Once $k$ surpasses $d$, redundant demonstrations fail to enhance reasoning—a phenomenon exacerbated in complex posets like $(\mathbb{N}, |)$, where dependencies form branching structures. This bottleneck underscores a fundamental limitation of ICL.
    \end{remark}

\subsection{Synthetic ICL Regression Experiment} %%ynthetic Regression-Based ICL Experiment
\label{appendix:icl-regression}

\paragraph{Motivation.} Inspired by the setup in \citet{guo2024how} and aligned with our theoretical framework, we implement a synthetic in-context learning (ICL) experiment for regression to directly observe the saturation behavior predicted by Theorem~\ref{thm:1}. This task provides a controlled, transparent environment to examine how the transformer’s representational capacity evolves with increasing demonstrations.

\paragraph{Problem Setup.} The task is a linear regression problem of the form:
\[
y = \langle \mathbf{w}, \mathbf{x} \rangle + \epsilon, 
\]
$\text{where } \mathbf{x} \in \mathbb{R}^d, \; \mathbf{w} \sim \mathcal{N}(0, I), \; \epsilon \sim \mathcal{N}(0, \sigma^2).$
We fix a weight vector $\mathbf{w} \in \mathbb{R}^{16}$ and generate batches of $(k+1)$ samples per task. Each task consists of $k$ in-context demonstration pairs $(\mathbf{x}_i, y_i)$ and a query point $(\mathbf{x}_q, y_q)$. The final point is treated as the prediction target.

\paragraph{Model Architecture.} We use a transformer encoder architecture with: \emph{(i) Input:} Concatenated $(\mathbf{x}, y)$ tokens of dimension $d+1$. \emph{(ii)} transformer layers, 4 attention heads, hidden size of 64. \emph{(iii)} The final token corresponds to the query, and its hidden state is projected via a linear layer to output the prediction $\hat{y}$.

\paragraph{Training.} The model is trained using MSE loss and Adam optimizer. We use a learning rate of $1e$-3 and StepLR scheduler with decay at 500 steps (factor 0.5). We train for 10,000 steps with a batch size of 128. Input noise is set to $\sigma = 0.05$ to mimic realistic data variance.

\paragraph{Evaluation Protocol.} After training, we fix the model and test its generalization across $k \in \{1, 2, \ldots, 32\}$. For each $k$, we run $2000$ trials and average the test MSE. This process is repeated across 5 random seeds to compute mean and standard deviation.

Notebook containing the complete training and evaluation code is available inside the same repository.

\paragraph{Results.} Figure~\ref{fig:icl_saturation} shows that model performance (measured in test MSE) improves sharply as $k$ increases up to $d = 16$, but saturates thereafter. This empirically confirms the rank constraint in Theorem~\ref{thm:1}, where the cumulative meta-update matrix $\Delta W_{\text{ICL}}$ has rank at most $\min(k, d)$. Beyond $k = d$, the transformer receives no representational gain, and additional examples yield diminishing returns.

\paragraph{Comparison to Related Work.} Our setup draws inspiration from \citet{guo2024how}, who demonstrate that transformers can perform ridge regression on representations via in-context updates. However, our experiment isolates this behavior in a minimal setting—where no representation function is required—allowing us to directly test the effect of linear independence among demonstrations and its alignment with theoretical limits on attention-based updates.

\section{Task Vector Analysis for ICL on Posets}
\label{appendix:task_vectors}
\subsection{Formal Setup and Definition}
Let $\widetilde{P}_{k,c}$ denote a prompt composed of:
\begin{itemize}
\itemsep-2pt
% \item $k$ demonstrations $\{(x_i, y_i)\}_{i=1}^{k}$, where $y_i \in R(x_i)$ for a poset relation $R$, 
% \subseteq \mathcal{X} \times \mathcal{X}$,
\item $k$ minimal demonstrations of format $(x_i \preceq y_i) \to \mathtt{True}$, if $x_i \preceq y_i$ or $(x_i \preceq y_i) \to \mathtt{False}$, if $y_i \preceq x_i$,
\item a query point of format $x_{k+1} \preceq y_{k+1}$ such that the query satisfies the $c$-complexity constraint: $(x_{k+1}, y_{k+1}) \notin \text{Hasse}(\widetilde{P}_{k})$, and either $x_{k+1}$ or $y_{k+1} \in \{k+1, \dots, k+c\}$.
\end{itemize}
We define the \emph{task vector} associated with prompt $\widetilde{P}_{k,c}$ as:
\[
\theta(\widetilde{P}_{k,c}) := \mathrm{Enc}_\ell(\widetilde{P}_{k,c})[t_{\rightarrow}] \in \mathbb{R}^d,
\]
where $\mathrm{Enc}_\ell(\cdot)$ is the embedding of the delimiter token at layer $\ell$ of the transformer. This vector is interpreted as the compressed representation of the relational structure implied by the prompt, consistent with the hypothesis class view $f(x; \theta)$ of ICL.

\subsection{Experimental Design}
We consider three task types:
\[
\mathcal{T} \in \left\{ (\N, <),\ (\{0,1\}^*, <),\ (\N, \mid) \right\},
\]
and let $\mathcal{D}_{\mathcal{T}}$ defines the collection of prompts for each $\mathcal{T}$. We evaluate the evolution of task vectors along two axes:
\begin{enumerate}
    \item \textbf{Varying Demonstrations $k$ (Fixed $c$)}: We fix complexity $c$ and compute
    \[
    \Theta^{\text{var-}k}_{\mathcal{T},c} = \left\{ \theta(\widetilde{P}_{k,c}^{(i)}) \right\}_{i=1}^{N}, \quad \text{where } \widetilde{P}_{k,c}^{(i)} \sim \mathcal{D}_{\mathcal{T}}.
    \]
    \item \textbf{Varying Complexity $c$ (Fixed $k$)}: We fix $k$ and compute
    \[
    \Theta^{\text{var-}c}_{\mathcal{T},k} = \left\{ \theta(\widetilde{P}_{k,c}^{(j)}) \right\}_{j=1}^{M}, \quad \text{where } \widetilde{P}_{k,c}^{(j)} \sim \mathcal{D}_{\mathcal{T}}.
    \]
\end{enumerate}
\subsection{t-SNE Visualization and Analysis}
To visualize the geometry of task vectors, we project $\Theta \subset \mathbb{R}^d \rightarrow \mathbb{R}^2$ using t-SNE (cosine distance, perplexity $10$). Separate plots are generated for:
\begin{itemize}\itemsep-2pt
    \item[i.] Each task $\mathcal{T}$,
    \item[ii.] Each variation type (demonstration or complexity),
\end{itemize}

\noindent
Each plot is annotated with the corresponding $k$ or $c$ value. We analyze:
\begin{itemize}\itemsep-2pt
    % \item[i.] Whether vectors form consistent clusters for fixed task identities,
    \item[i] Whether increasing $k$ or $c$ induces representational drift or collapse,
    \item[ii.] Whether saturation occurs, consistent with theoretical rank bounds (Theorem~\ref{thm:1}).
\end{itemize}

\subsubsection{Connection to Meta-Optimization Limits}

The representational capacity of ICL under the attention-as-optimization hypothesis using Theorem \ref{thm:1} is constrained by the following:
% \begin{theorem}[Rank Bound of ICL Updates]
% Let $\Delta W_{\mathrm{ICL}} = \sum_{i=1}^{k} (W_V x_i') \otimes (W_K x_i')$. Then
\[
\mathrm{rank}(\Delta W_{\mathrm{ICL}}) \leq \min(k, d),
\]
where $d$ is the embedding dimension and $x_i'$ are the embeddings of the demonstration tokens. Hence, for $k > d$, additional demonstrations yield no representational gain.
% \end{theorem}

This explains empirical saturation: if task vectors $\theta(\widetilde{P}_{k,c})$ collapse to a low-rank subspace for $k > d$ or for large $c$, then performance plateau and geometric flattening in t-SNE should co-occur.

\subsection{Supplemental Experimental Results on Task Vectors}
\label{appendix:tSNEwithRelDesc}
We have followed \citeauthor{hendel2023context}'s way for finding the layer with maximum accuracy within a predefined range to plot the t-SNE of the task vectors. Along the line, the 2D t-SNE plots have been initialized with $41$ random states. All the experiments have been conducted on a 48 GB NVIDIA RTX A6000 GPU.

\subsubsection{Task Vector Geometry under varying Demonstrations with \textsf{Pythia-2.8B}}
\begin{figure}[!ht]
    \centering
    \captionsetup[subfigure]{font=small, skip=2pt}
    \begin{subfigure}[b]{0.32\textwidth}
        \centering
        \includegraphics[width=\linewidth]{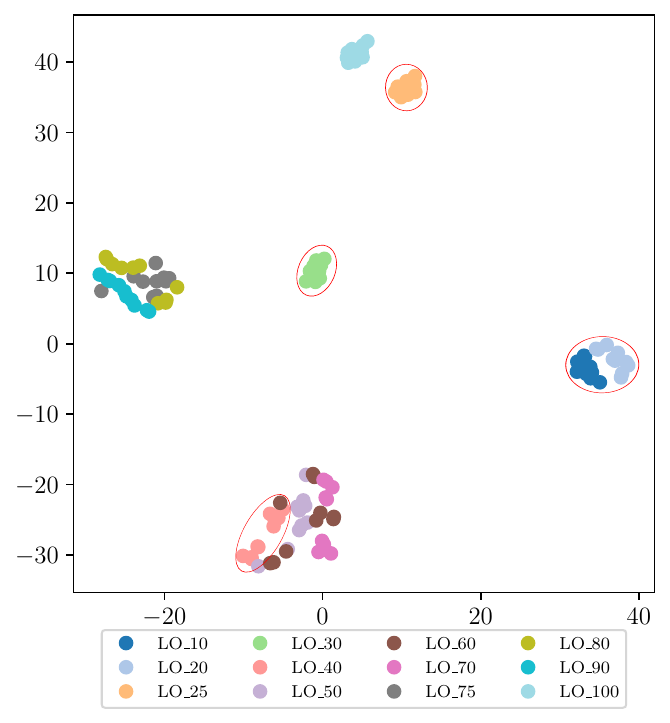}
        \caption{\texttt{LO} ($(\N,<)$)}
        \label{fig:8a}
    \end{subfigure}
    \hfill
    \begin{subfigure}[b]{0.32\textwidth}
        \centering
        \includegraphics[width=\linewidth]{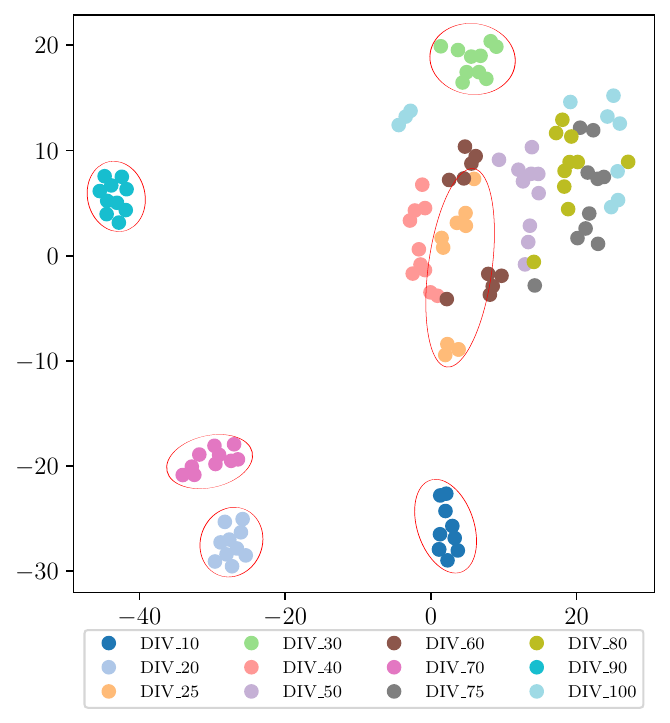}
        \caption{\texttt{DIV} ($(\N,\mid)$)}
        \label{fig:8b}
    \end{subfigure}
    \hfill
    \begin{subfigure}[b]{0.32\textwidth}
        \centering
        \includegraphics[width=\linewidth]{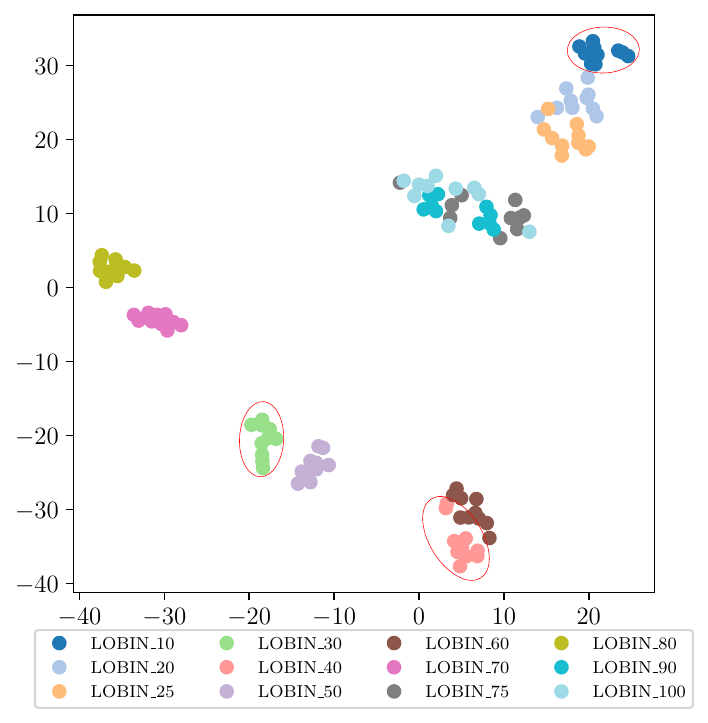}
        \caption{\texttt{LOBIN} ($(\{0,1\}^*, <)$)}
        \label{fig:8c}
    \end{subfigure}

    \caption{t-SNE projections of task vectors from \textsf{Pythia-2.8B} under increasing number of demonstrations ($k \in [10, 100]$) with fixed complexity. Subfigures \ref{fig:8a}-\ref{fig:8c} correspond to \texttt{LO}, \texttt{DIV}, and \texttt{LOBIN}, respectively. Low-shot clusters are well-separated, with convergence and overlap appearing as $k$ increases, reflecting latent saturation.}
    \label{fig:tsne-pythia-shots}
\end{figure}

To test the generality of latent representation trends across model families, we extend our analysis to the \textsf{Pythia-2.8B} model—part of an open suite of decoder-only architectures designed for transparent scaling and training behavior. Following the setup above we fix the complexity and vary the number of in-context demonstrations $k \in \{10, 20, \ldots, 100\}$ across three tasks: \texttt{DIV} ($(\N, \mid)$), \texttt{LO} ($(\N, <)$), and \texttt{LOBIN} ($(\{0,1\}^*, <)$). For each $k$, we generate ten prompts using the same demonstrations with varying queries, extract task vectors from a fixed transformer layer, and visualize them using t-SNE.

As shown in Figure~\ref{fig:tsne-pythia-shots}, Pythia exhibits representational trends broadly consistent with our \textsf{Llama3} findings (Fig. \ref{fig:tsne-shots}). In the low-shot regime ($k \leq 30$), clusters are well-separated, indicating distinct task representations. As $k$ increases, task vector clusters progressively collapse, converging toward shared latent regions indicative of representational saturation. This pattern confirms the broader validity of the saturation phenomenon under ICL.

Notably, \textsf{Pythia-2.8B} shows a relatively slower convergence in task vector geometry compared to \textsf{Llama3}, particularly for \texttt{DIV} and \texttt{LOBIN}. This may reflect Pythia’s architectural emphasis on scaling consistency rather than peak performance, as intended by its design \citep{biderman2023pythia}. For instance, \texttt{DIV} clusters remain spread even at $k = 70$, suggesting limited generalization over deeper DAG relations. In contrast, \texttt{LO} shows earlier saturation with cluster overlap emerging from $k = 40$, highlighting the lower representational burden of total orders. The \texttt{LOBIN} task retains intermediate separation, echoing patterns observed in \textsf{Llama3}.

These observations visually affirm Theorem~\ref{thm:1}, which bounds ICL's representational capacity by the model’s embedding rank. The consistent saturation across both Pythia and Llama families underscores this limitation and motivates future directions in attention mechanisms and model design for relational tasks.

\subsubsection{Task Vector Geometry under varying Complexity with \textsf{Pythia-2.8B}}
\label{ssec:tvpythiacom}

\begin{figure}[!ht]
    \centering
    \captionsetup[subfigure]{font=small, skip=2pt}

    \begin{subfigure}[b]{0.32\textwidth}
        \centering
        \includegraphics[width=\linewidth]{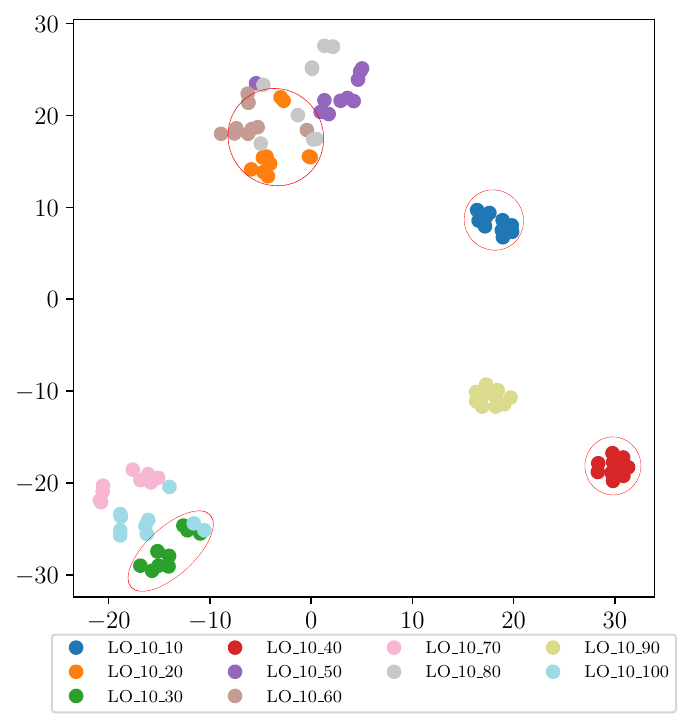}
        \caption{\texttt{LO} ($(\N,<)$)}
    \end{subfigure}
    \hfill
    \begin{subfigure}[b]{0.32\textwidth}
        \centering
        \includegraphics[width=\linewidth]{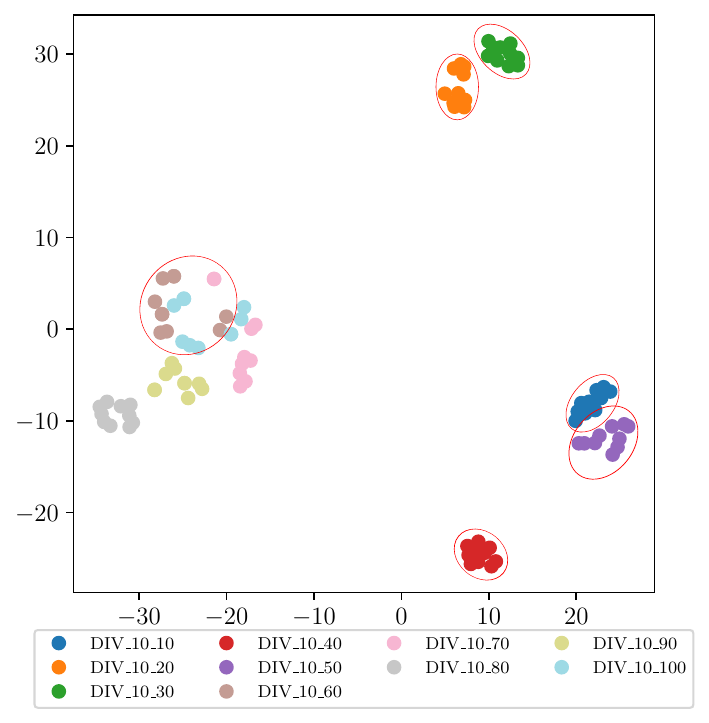}
        \caption{\texttt{DIV} ($(\N,\mid)$)}
    \end{subfigure}
    \hfill
    \begin{subfigure}[b]{0.32\textwidth}
        \centering
        \includegraphics[width=\linewidth]{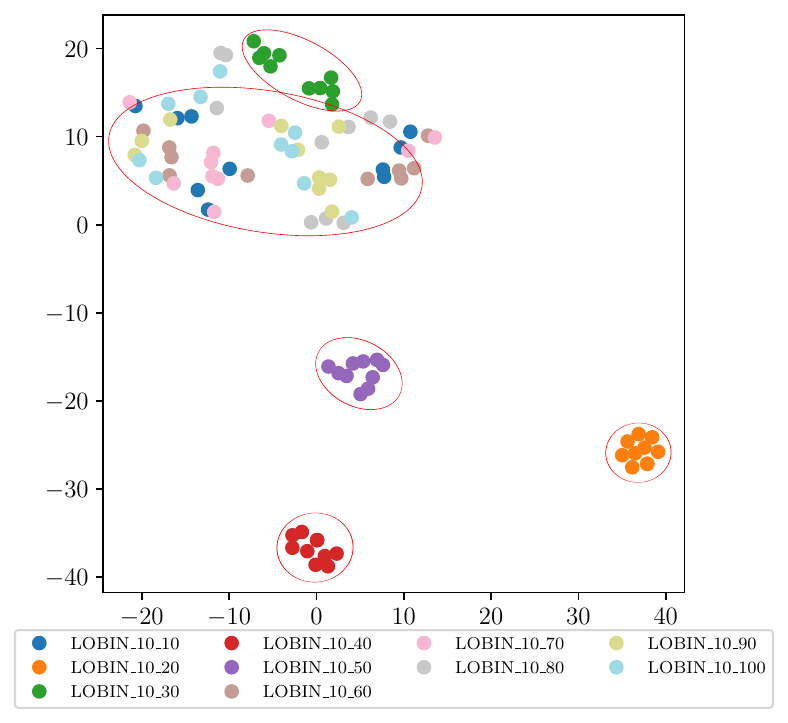}
        \caption{\texttt{LOBIN} ($(\{0,1\}^*, <)$)}
    \end{subfigure}

    \caption{t-SNE projections of task vectors from \textsf{Pythia-2.8B} under increasing complexity ($c \in [10, 100]$) with fixed demonstrations ($k = 10$). Each subfigure corresponds to a task: (a) \texttt{LO}, (b) \texttt{DIV}, and (c) \texttt{LOBIN}. Cluster convergence at higher $c$ levels illustrates latent saturation as predicted by Theorem~\ref{thm:1}.}
    \label{fig:tsne-pythia-complexity}
\end{figure}

In this setup we consider the prompts $\widetilde{P}_{k,c}$ along with the instruction $I$, that is restricting $|P_{k,c}[T^\prime]|$ to one. To assess how the complexity impacts latent representations in ICL, we evaluate the \textsf{Pythia-2.8B} model under fixed-shot prompts ($k = 10$) while varying the complexity of the target query $c \in \{10, 20, \dots, 100\}$. Following the protocol as discussed above, we generate ten prompts for each complexity level by pairing a constant set of demonstrations with increasingly complex queries. We extract task vectors from a fixed transformer layer and visualize their geometry via t-SNE.

As shown in Figure~\ref{fig:tsne-pythia-complexity}, \textsf{Pythia-2.8B} exhibits consistent cluster convergence across increasing $c$, though the saturation dynamics vary by task. For \texttt{LO} ($(\N, <)$), task vector clusters collapse rapidly by $c = 50$, consistent with the linear structure's limited inferential depth. In contrast, \texttt{DIV} ($(\N, \mid)$) shows more persistent separation, with full convergence only emerging beyond $c = 70$. \texttt{LOBIN} ($(\{0,1\}^*, <)$) occupies an intermediate regime—maintaining moderate separation through mid-range complexities and gradually collapsing at higher levels ($c \geq 80$).

%%%%%%%%%%%%%%%%%%%%%%%%%%%%%%%%%%%%%%%%%%%%%%%%%%%%%%%%%%%%%%%%%%%%%%%%%%%%%%%%%%%
\begin{figure*}[!ht]
    % \vspace*{-3mm}
    \captionsetup[subfigure]{font=small, skip=2pt}
    \begin{subfigure}[b]{0.32\textwidth}
        \includegraphics[trim={2mm 0 0 0}, clip, width=\textwidth]{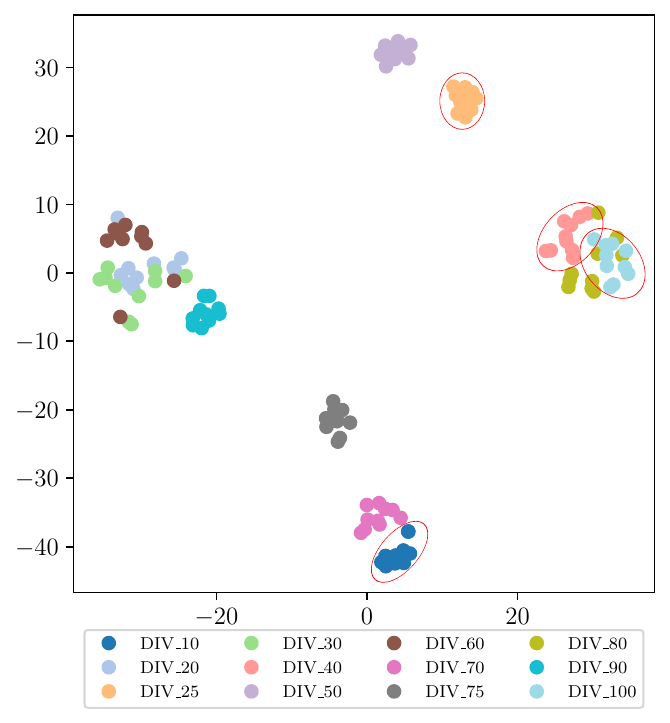}
        \caption{\texttt{DIV}($(\N, \mid)$)}
        \label{fig:9a}
    \end{subfigure}\hfill
   %  \begin{subfloatrow}
   %    \fcapside[\FBwidth]
   %      {\caption{}\label{fig:9b}}%
   %      {\includegraphics[width=0.45\textwidth]{./images/Pythia_tsne_LO_Shot_L_Ideal_hl.pdf}}%
   % \end{subfloatrow}
   %  \begin{subfloatrow}
   %    \fcapside[\FBwidth]
   %      {\caption{}\label{fig:9c}}%
   %      {\includegraphics[trim={2mm 0 0 0}, clip, width=0.45\textwidth]{./images/Pythia_tsne_LOBIN_Shot_L_hl.pdf}}%
   % \end{subfloatrow}
   %  \begin{subfloatrow}
   %    \fcapside[\FBwidth]
   %      {\caption{}\label{fig:9d}}%
   %      {\includegraphics[width=0.45\textwidth]{./images/tsne_DIV_Comp_L_17.05.2025_20_43_33_hl.pdf}}% ./images/tsne_DIV_Comp_S_18.05.2025_12_04_43.pdf
   % \end{subfloatrow}
    \begin{subfigure}[b]{0.32\textwidth}
        \includegraphics[trim={2mm 0 0 0}, clip, width=\textwidth]{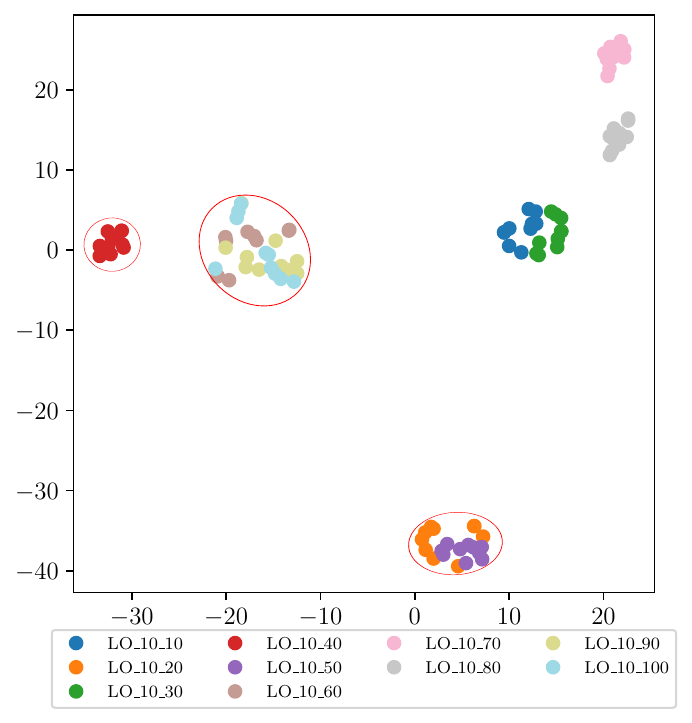}
        \caption{\texttt{LO}($(\N, <)$)}
        \label{fig:9e}
    \end{subfigure}\hfill
   \begin{subfigure}[b]{0.32\textwidth}
        \includegraphics[trim={2mm 0 2mm 0mm}, clip, width=\textwidth]{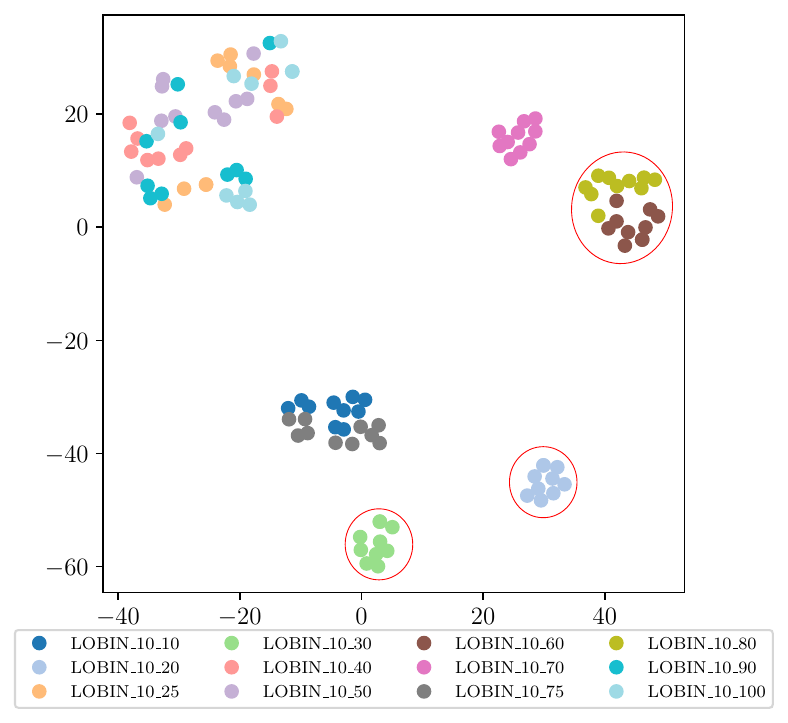}
        \caption{\texttt{LOBIN}($(\{0,1\}^*, <)$)}
        \label{fig:9f}
    \end{subfigure}
    \caption{Subfigure \ref{fig:9a} plots the t-SNE projections of task vectors under increasing number of demonstrations ($k\in[10, 100]$) with fixed complexity ($c=10$) for the poset \texttt{DIV}($(\N, \mid)$). Subfigures \ref{fig:9e} \& \ref{fig:9f} plot the same under increasing complexities ($c\in[10, 100]$) with fixed demonstrations ($k=10$) for the posets \texttt{LO}($(\N, <)$) and \texttt{LOBIN}($(\{0,1\}^*, <)$) respectively. Note that, these plots represent the task vectors for the latent representation of the model \textsf{Llama3} where the prompts $\widetilde{P}_{k,c}$ has been augmented with the instruction $I$ as mentioned in Section \ref{ssec:tvpythiacom}.}
    \label{fig:llamareldesc}
\end{figure*}
%%%%%%%%%%%%%%%%%%%%%%%%%%%%%%%%%%%%%%%%%%%%%%%%%%%%%%%%%%%%%%%%%%%%%%%%%%%%%%%%%%%

Compared to \textsf{Llama3} (in Fig. \ref{fig:tsne-shots}), \textsf{Pythia}'s latent space appears more diffuse for structurally intricate tasks such as \texttt{DIV}. This aligns with Pythia's design motivation as outlined in \citet{biderman2023pythia}, where training stability and scaling transparency are prioritized over optimization for downstream accuracy. These geometric observations reinforce our theoretical result (Theorem~\ref{thm:1}) that ICL updates are rank-bounded by the embedding dimension, with saturation manifesting visibly as latent collapse in t-SNE space.

Together with the earlier shot-based findings, this analysis highlights that decoder-based LLMs consistently struggle to encode deep relational abstractions through ICL alone. Models with improved inductive generalization will require architectural changes to mitigate the geometric bottlenecks revealed here.

Figure \ref{fig:llamareldesc} illustrates analogous study conducted on \textsf{Llama3} under similar setup.

\section{Additional Experimental Results}

\begin{figure}[!ht]
    \centering
    \begin{subfigure}{0.32\textwidth}
    \centering
        \includegraphics[trim={0 0 0 0mm}, clip, width=\textwidth]{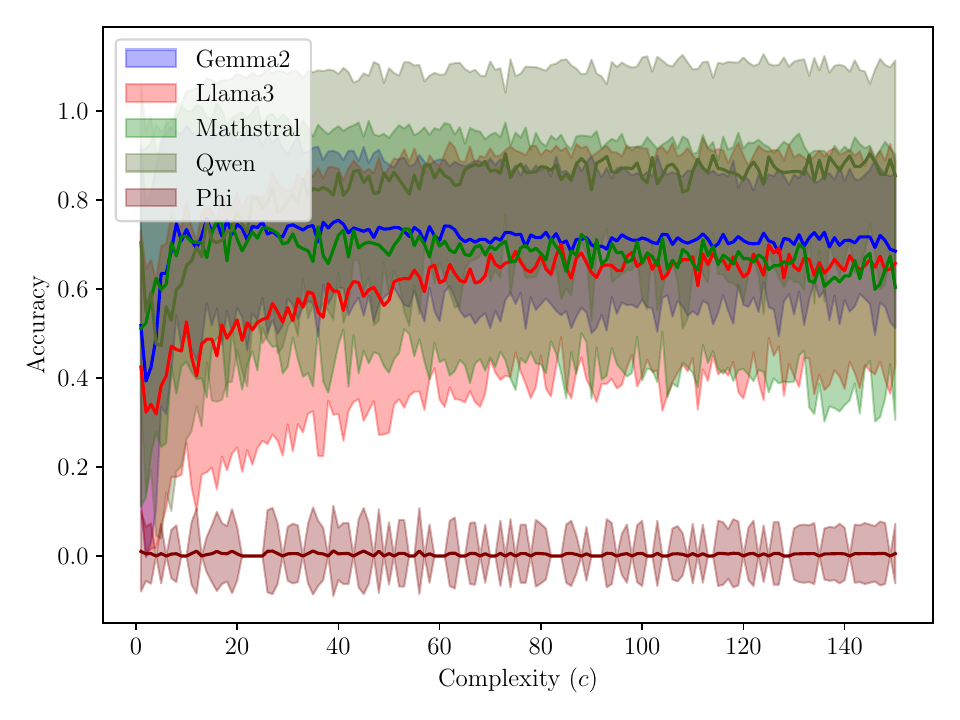}\vspace*{-10pt}
        \caption{}
        \label{fig:adda}
    \end{subfigure}\hspace{\fill}
    \begin{subfigure}{0.32\textwidth}
    \centering
        \includegraphics[trim={0 0 0 0mm}, clip, width=\textwidth]{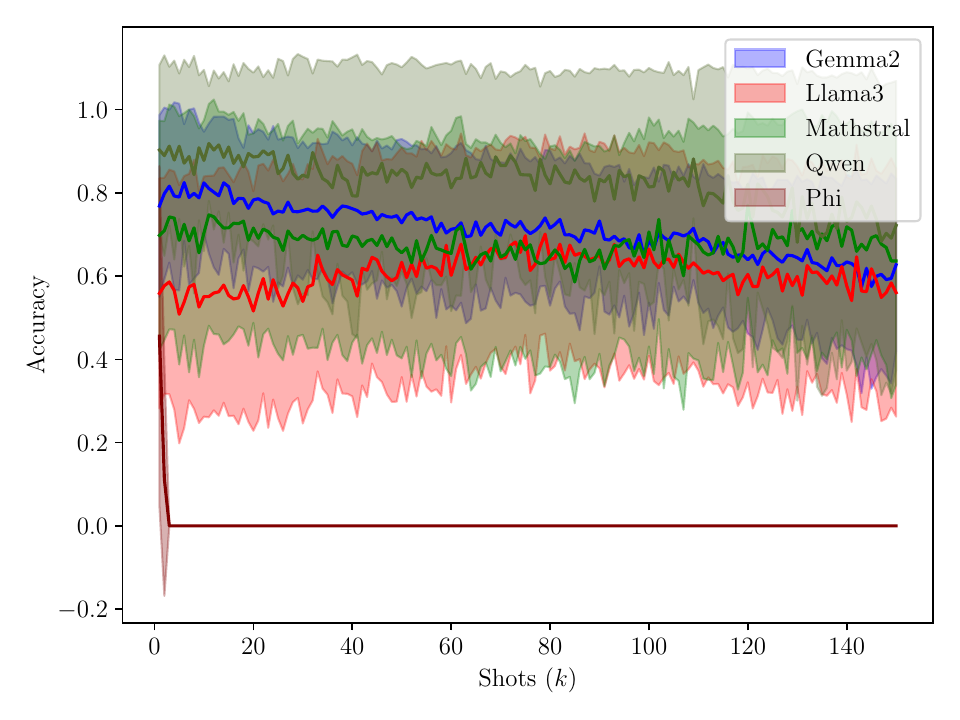}\vspace*{-10pt}
        \caption{}
        \label{fig:addb}
    \end{subfigure}\hspace{\fill}
    \begin{subfigure}{0.32\textwidth}
    \centering
        \includegraphics[trim={0 0 0 0mm}, clip, width=\textwidth]{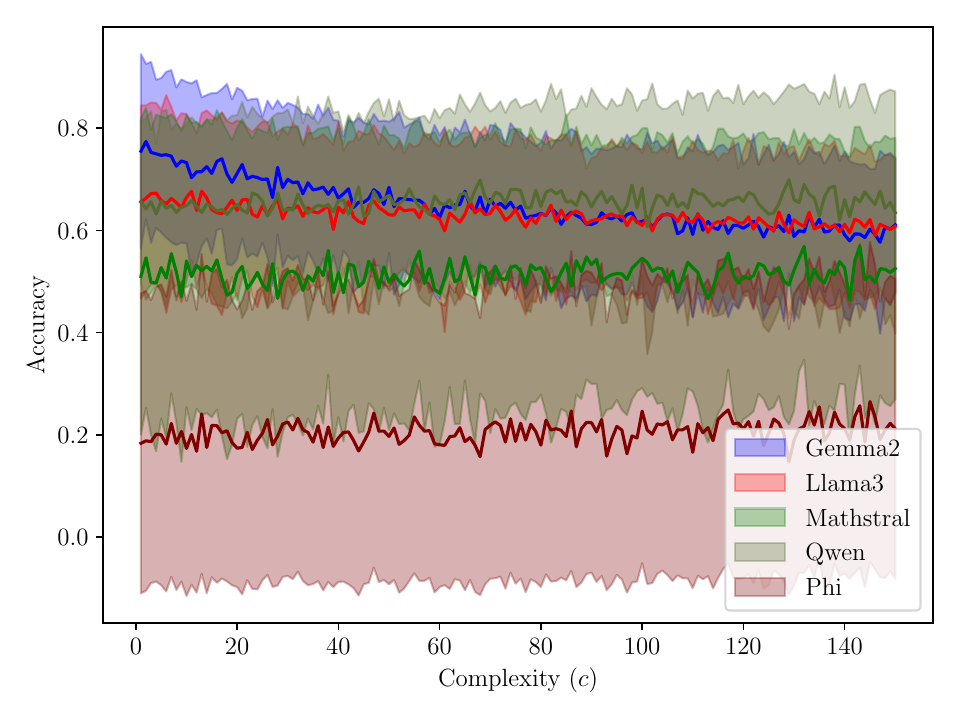}\vspace*{-10pt}
        \caption{}
        \label{fig:addc}
    \end{subfigure}\\
    \begin{subfigure}{0.32\textwidth}
    \centering
        \includegraphics[trim={0 0 0 0mm}, clip, width=\textwidth]{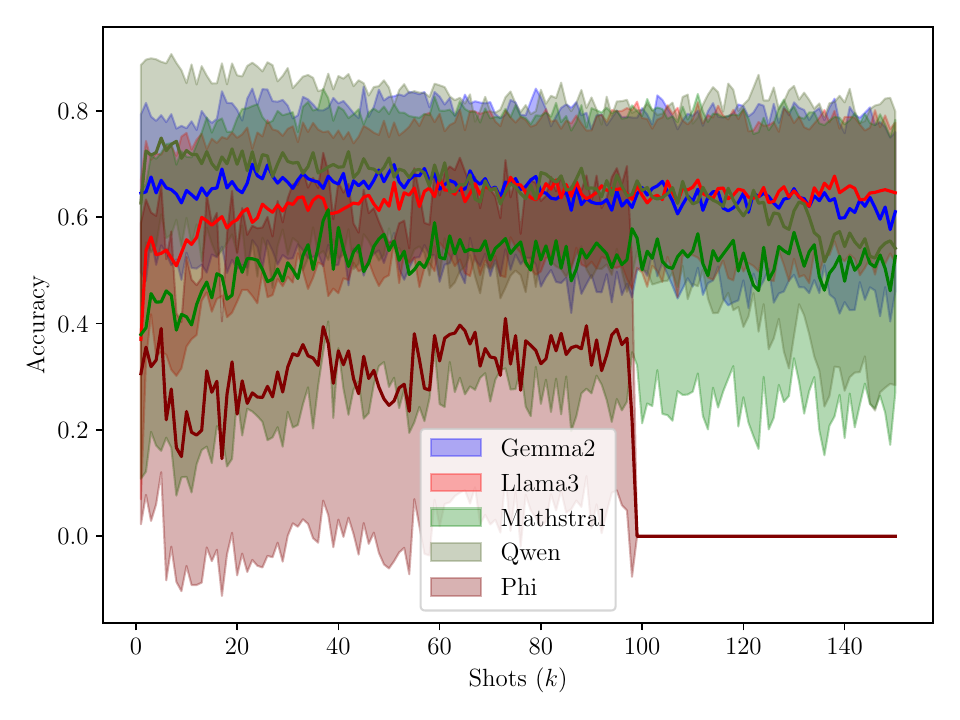}\vspace*{-10pt}
        \caption{}
        \label{fig:addd}
    \end{subfigure}\hspace{\fill}
    \begin{subfigure}{0.32\textwidth}
    \centering
        \includegraphics[trim={0 0 0 0mm}, clip, width=\textwidth]{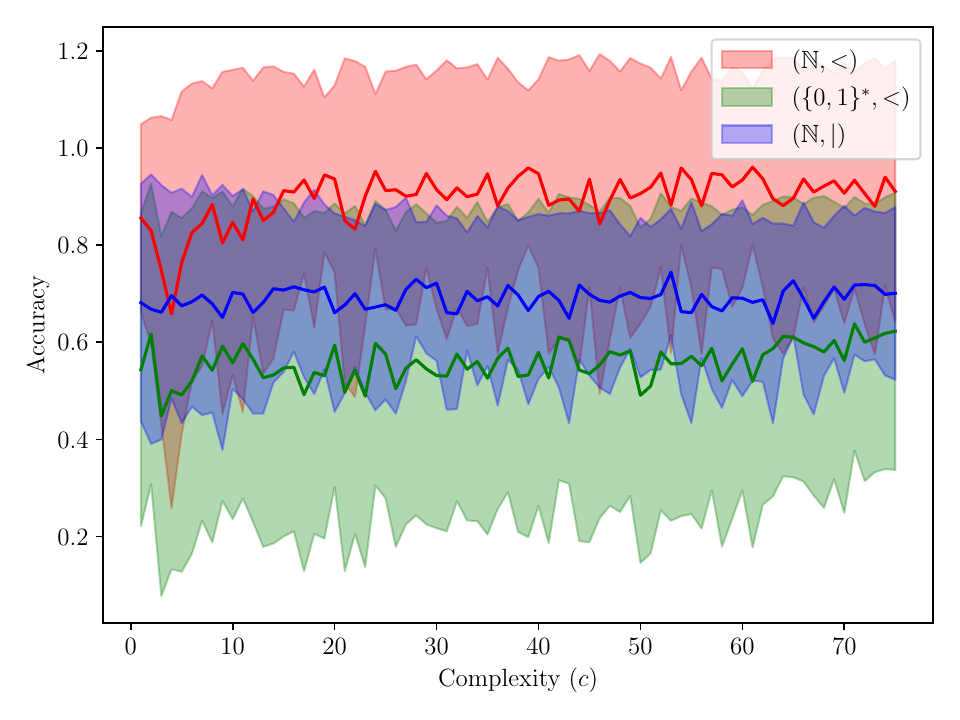}\vspace*{-10pt}
        \caption{}
        \label{fig:adde}
    \end{subfigure}  \hspace{\fill}
    \begin{subfigure}{0.32\textwidth}
    \centering
        \includegraphics[trim={0 0 0 0mm}, clip, width=\textwidth]{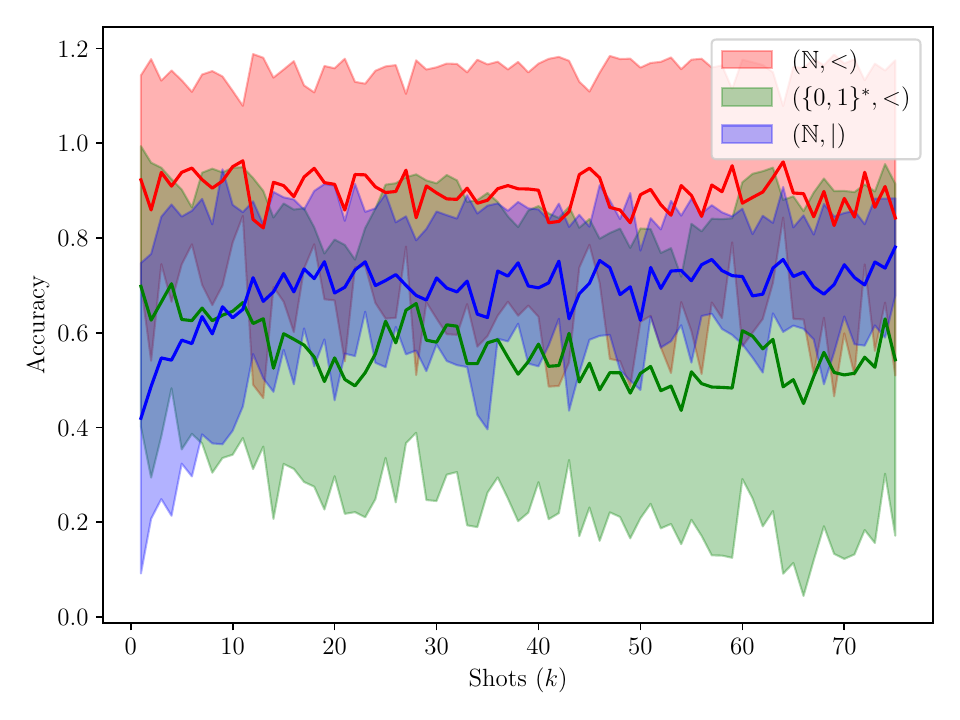}\vspace*{-10pt}
        \caption{}
        \label{fig:addf}
    \end{subfigure}\\
    \begin{subfigure}{0.32\textwidth}
    \centering
        \includegraphics[trim={0 0 0 0mm}, clip, width=\textwidth]{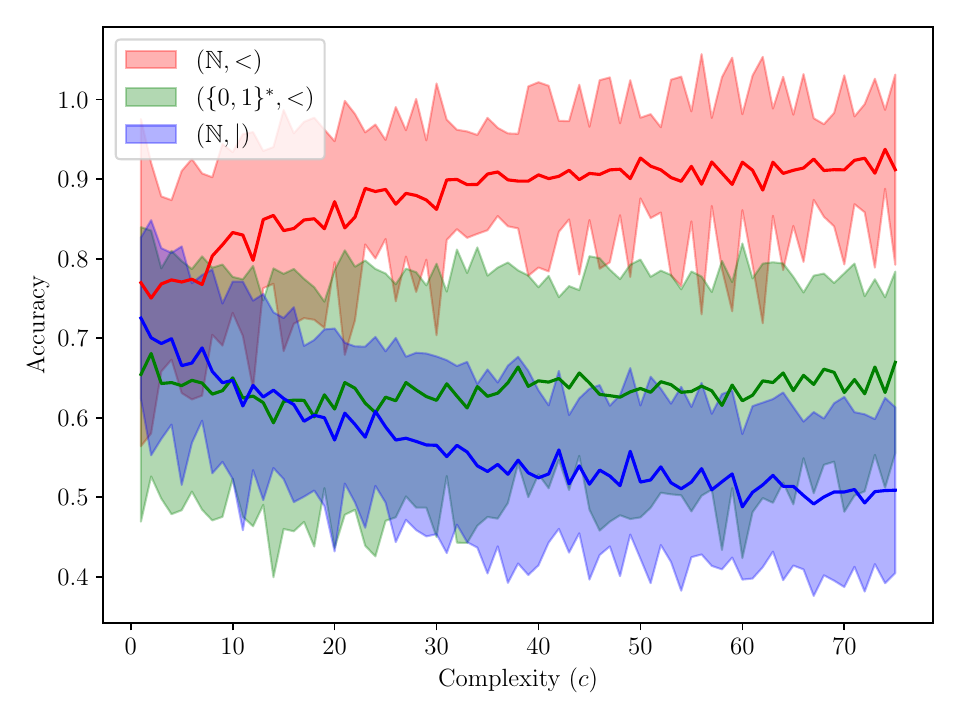}\vspace*{-10pt}
        \caption{}
        \label{fig:addg}
    \end{subfigure}
    \begin{subfigure}{0.32\textwidth}
    \centering
        \includegraphics[trim={0 0 0 0mm}, clip, width=\textwidth]{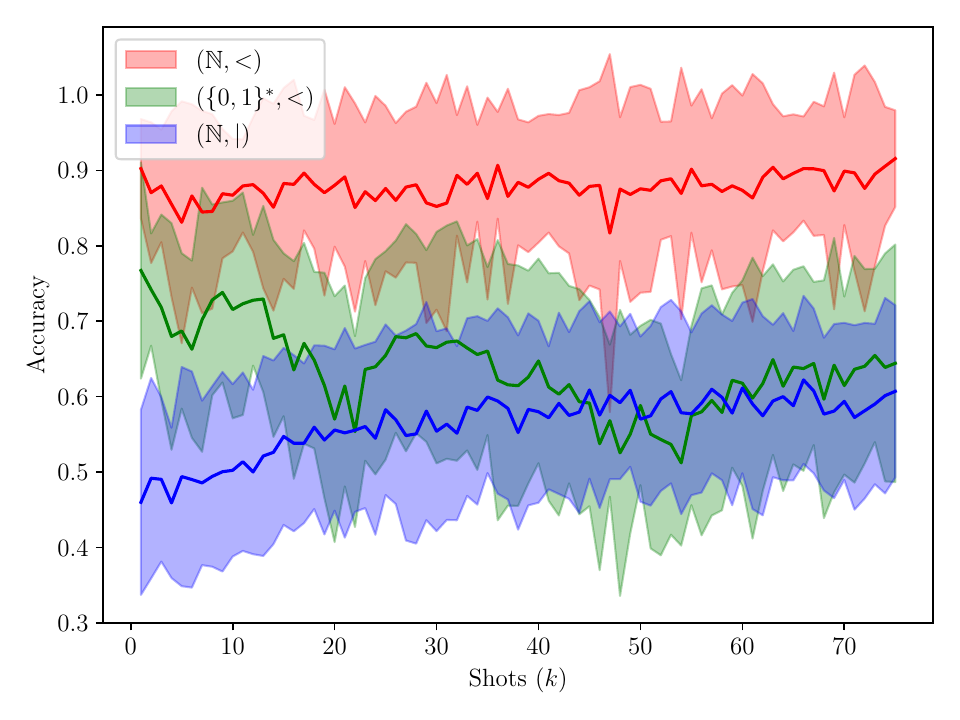}\vspace*{-10pt}
        \caption{}
        \label{fig:addh}
    \end{subfigure}
    \caption{Subfigure \ref{fig:adda} describes the mean accuracy plot for the sequence $\set{\L(P_{k,c})}_{k =\set{1, 2, \ldots, 150}}$ against varying complexities and subfigure \ref{fig:addb} describes the same evaluation scheme on $\set{\L(P_{k,c})}_{c =\set{1, 2, \ldots, 150}}$ against varying shots for the poset $(\N, <)$. Subfigures \ref{fig:addc} and \ref{fig:addd} illustrate the similar experimental result on poset $(\N, \mid)$. Figure \ref{fig:adde} and \ref{fig:addf} depict mean accuracy plot for the sequence $\set{\L(P_{k,c})}_{k =\set{1, 2, \ldots, 150}}$ against varying complexities and the same metric on $\set{\L(P_{k,c})}_{c =\set{1, 2, \ldots, 150}}$ against varying shots for \textsf{GPT-3.5-Turbo} respectively. Similarly, figure \ref{fig:addg} and \ref{fig:addh} illustrates the same for \textsf{GPT-4.o-mini}.}
    \label{fig_additional_example}
\end{figure}

In Figure \ref{fig_additional_example} we provide a few diagrams representing the effectiveness of ICL through mean accuracy plots across various cases.
Before closing the analysis of our empirical studies, here we present how well the language models correspond with each other under the regime of minimal prompts for partially ordered set. Because of the involvement of more than two objects to be ranked, we employ Kendall's $W: \N^{m\times n} \to [0,1]$, a non-parametric statistic measuring ranking correlation between $m$ judges (the language models) and $n$ items (the tasks on which the language models have been evaluated) \citet{kendall1939problem}.
To rank the items, the squared deviation between the sum of ranks of different judges ($R_i = \sum_{j=1}^m r_{ij}$) and their mean value is usually calculated and the statistic is expressed by \[W = \frac{\sum_{i=1}^n(R_i - \bar{R})^2}{\frac{m^2}{12}(n^3-n)}.\] For our case, the ranking was done based on the mean cumulative accuracy metric. Now if, the judges have weights assigned to them, which is the number of parameter in our case, the expression for Kendall's $W : \mathbb{\left({R}^+\right)}^{m\times n} \to [0,1]$ becomes: \[W = \frac{\sum_{i=1}^n(R_i - \bar{R})^2}{\frac{n}{12}(n^2-1)},\] where $R_i = \sum_{j=1}^m w_jr_{ij}$ and the weight $w_j$ assigned to judge $j$ is normalized. The higher the value of Kendall's W, the closer the models towards its ICL performance. Table \ref{tab:Kendall} demonstrates the behavior of the LLMs on tasks employed.
\begin{table}[ht]
        \centering
        \caption{Ranks and Calculation of Kendall's W showing non-trivial correspondence between the language models.}
        \begin{tabular}{ll}
            \hline
            Judges & Rank of Items\\\hline
             Gemma & $(\N, <), (\N, |), (\set{0,1}^*, <)$\\
             Lamma & $(\N, |), (\N, <), (\set{0,1}^*, <)$\\
             Mathstral & $(\N, <), (\N, |), (\set{0,1}^*, <)$\\
             Qwen2.5-math & $(\N, <), (\N, |), (\set{0,1}^*, <)$\\
             Phi & $(\N, |), (\N, <), (\set{0,1}^*, <)$\\\hline
             Kendall's W & $0.7506$\ (medium)\\\hline
        \end{tabular}
        \label{tab:Kendall}
\end{table}

% \begin{figure}
%     \centering
%     \begin{subfigure}{0.24\textwidth}
%         \centering
%         \includegraphics[width=\textwidth, trim={0 0 0 7mm}, clip]{V1/images/LO10AlongRowPointwise.pdf}
%         \caption{}
%         \end{subfigure}
%     \begin{subfigure}{0.24\textwidth}
%         \centering
%         \includegraphics[trim={0 0 0 7mm}, clip, width=\textwidth]{V1/images/LO10AlongColPointwise.pdf}
%         \caption{}
%     \end{subfigure}
% % \end{figure}
% % \begin{figure}
% % \centering
%     \begin{subfigure}{0.24\textwidth}
%         \centering
%         \includegraphics[trim={0 0 0 7mm}, clip, width=\textwidth]{V1/images/DIV10AlongRowPointwise.pdf}
%         \caption{}
%     \end{subfigure}
%     \begin{subfigure}{0.24\textwidth}
%         \centering
%         \includegraphics[trim={0 0 0 7mm}, clip, width=\textwidth]{V1/images/DIV10AlongColPointwise.pdf}
%         \caption{}
%     \end{subfigure}
%     \caption{Subfigure (a) describes the mean accuracy plot for the sequence $\set{\L(P_{k,c})}_{k =\set{1, 2, \ldots, 150}}$ against varying complexities and subfigure (b) describes the same evaluation scheme on $\set{\L(P_{k,c})}_{c =\set{1, 2, \ldots, 150}}$ against varying shots for the poset $(\N, <)$. Subfigure (a) and (b) illustrate the similar experimental result on poset $(\N, \mid)$.}
% \end{figure}

\end{document}